\documentclass[]{fairmeta}

\usepackage{hyperref}
\usepackage{url}
\usepackage{amsmath}
\usepackage{amsfonts} 
\usepackage{amssymb}
\usepackage{xargs}                      
\usepackage{svg}
\usepackage{tablefootnote}
\usepackage{graphicx}
\usepackage{wrapfig}
\usepackage{caption}
\usepackage{subcaption}
\usepackage{changepage}

\usepackage{appendix}
\usepackage{pdflscape} 
\usepackage{color}
\usepackage{float}

\title{On the origin of neural scaling laws: \\ from random graphs to natural language}

\author[1,2,*]{Maissam Barkeshli}
\author[3,\dagger, *]{Alberto Alfarano}
\author[1]{Andrey Gromov}

\affiliation[1]{Meta Superintelligence Labs, FAIR}
\affiliation[2]{Department of Physics, University of Maryland, College Park and Joint Quantum Institute}
\affiliation[3]{Axiom Math}

\contribution[*]{Equal contribution}
\contribution[\dagger]{Work done at Meta}

\abstract{
Scaling laws have played a major role in the modern AI revolution, providing practitioners predictive power over how the model performance will improve with increasing
data, compute, and number of model parameters. This has spurred an intense interest in the origin of neural scaling laws, with a common suggestion being that they arise from power law structure already present in the data. In this paper we study scaling laws for transformers trained to predict random walks (bigrams) on graphs with tunable complexity. We demonstrate that this simplified setting already gives rise to neural scaling laws even in the absence of power law structure in the data correlations. We further consider dialing down the complexity of natural language systematically, by training on sequences sampled from increasingly simplified generative language models, from 4,2,1-layer transformer language models down to language bigrams, revealing a monotonic evolution of the scaling exponents. Our results also include scaling laws obtained from training on random walks on random graphs drawn from Erdös-Renyi and scale-free Barabási-Albert ensembles. Finally, we revisit conventional scaling laws for language modeling, demonstrating that several essential results can be reproduced using 2 layer transformers with context length of 100, provide a critical analysis of various fits used in prior literature, demonstrate an alternative method for obtaining compute optimal curves as compared with current practice in published literature, and provide preliminary evidence that maximal update parameterization may be more parameter efficient than standard parameterization. 
}

\begin{document}

\maketitle

\section{Introduction}

One of the most important lessons in modern deep learning is the steady improvement in model capabilities as additional compute resources and data are effectively leveraged \citep{sutton2019bitter}. This was partially quantified through the characterization of neural scaling laws \citep{cortes1993learning,hestness2017deep,kaplan2020scaling,henighan2020scaling,hoffmann2022training}, which demonstrate that across many vision and natural language tasks, the test loss decreases predictably as a simple power law over many orders of magnitude in number of model parameters $N$, dataset size $D$, and amount of compute $C$. The discovery of neural scaling laws has had significant impact in practice for language model pretraining. It allows practitioners to determine how to optimally scale the model size and dataset size with compute \citep{kaplan2020scaling,hoffmann2022training,chowdhery2022palm,grattafiori2024llama3,yang2024qwen2,yang2025qwen25,liu2024deepseekv2,jiang2024mixtral,tian2025moeScaling}. It also provides a way to benchmark algorithmic breakthroughs in architectures, optimization, and data. 

These empirical results have led to significant theoretical work in trying to understand the origin of neural scaling laws. Specifically, why is there a power law decrease in the test loss over many orders of magnitude in $N$, $D$, and $C$, and what sets the exponents of the power laws? A clear answer to this question may be of significant practical value, since if we understand what sets the exponents, we might understand the extent to which they can be increased, thus increasing the asymptotic efficiency of deep learning methods.

A popular suggestion has been that the power law scaling in the test loss originates from power laws that are already present in the dataset itself. For example, it is well-known that the frequency of words in a corpus of text follows Zipf's law, with many other power laws having also been characterized in natural language corpora \citep{piantadosi2014zipf,altmann2016statistical}. Natural images also exhibit power laws  in their spectra \citep{ruderman1994statistics,maloney2022solvable}. Many theoretical works have shown that in linear or kernel regression, power laws in the test loss do in fact originate from power laws in the data (or in features defined in terms of the data) \citep{bordelon2020,bahri2021explaining,Spigler_2020,maloney2022solvable,lin2024scaling,paquette2024fourplus3,bordelon2024dynamical}. More generally, if we assume that models need to learn a discrete set of tasks to achieve a particular value of test loss, and if these tasks are distributed with power law weighting, then a power law in the test loss follows \citep{michaud2024quantizationmodelneuralscaling,ren2025emergence}. 

The above theories based on linear models with mean square error (MSE) loss are rather far from the setting of auto-regressive sequence modeling with cross-entropy loss. Consequently it is not clear to what extent they are representative of the neural scaling laws seen in natural language modeling. A potentially fruitful approach would be to study sequence modeling with datasets of \it tunable complexity\rm, where one can systematically dial down from a realistic limit to a highly simplified limit, and track the change in the neural scaling law behavior. An additional benefit of datasets with tunable complexity is that they could potentially allow more appropriate comparisons of models across different scales: a proper comparison of a small model to a large model should also appropriately increase the complexity of the dataset. Complexity of the dataset in turn is not measured simply in terms of size of the dataset, but also in terms of the degree of correlations, and hierarchical, compositional structure \citep{cagnetta2024deep}. 

To this end, we propose to study sequence modeling of random walks on graphs. Graphs and their generalizations -- hypergraphs and hierarchical graphical structures -- can capture a large portion of many kinds of data of interest, including many features of language, games, and formal mathematics. In particular, walks on graphs can correspond to generating from $n$-gram models (Sec. \ref{sec:ngrams}), providing a simplified model of language. On the other hand, at a more abstract level, walks on graphs can be used to model stepwise inference and chain of thought reasoning \citep{khona2024understandingstepwiseinferencetransformers,Besta_2024}. 

\begin{figure}[t]
    \centering
    \includegraphics[width=\linewidth]{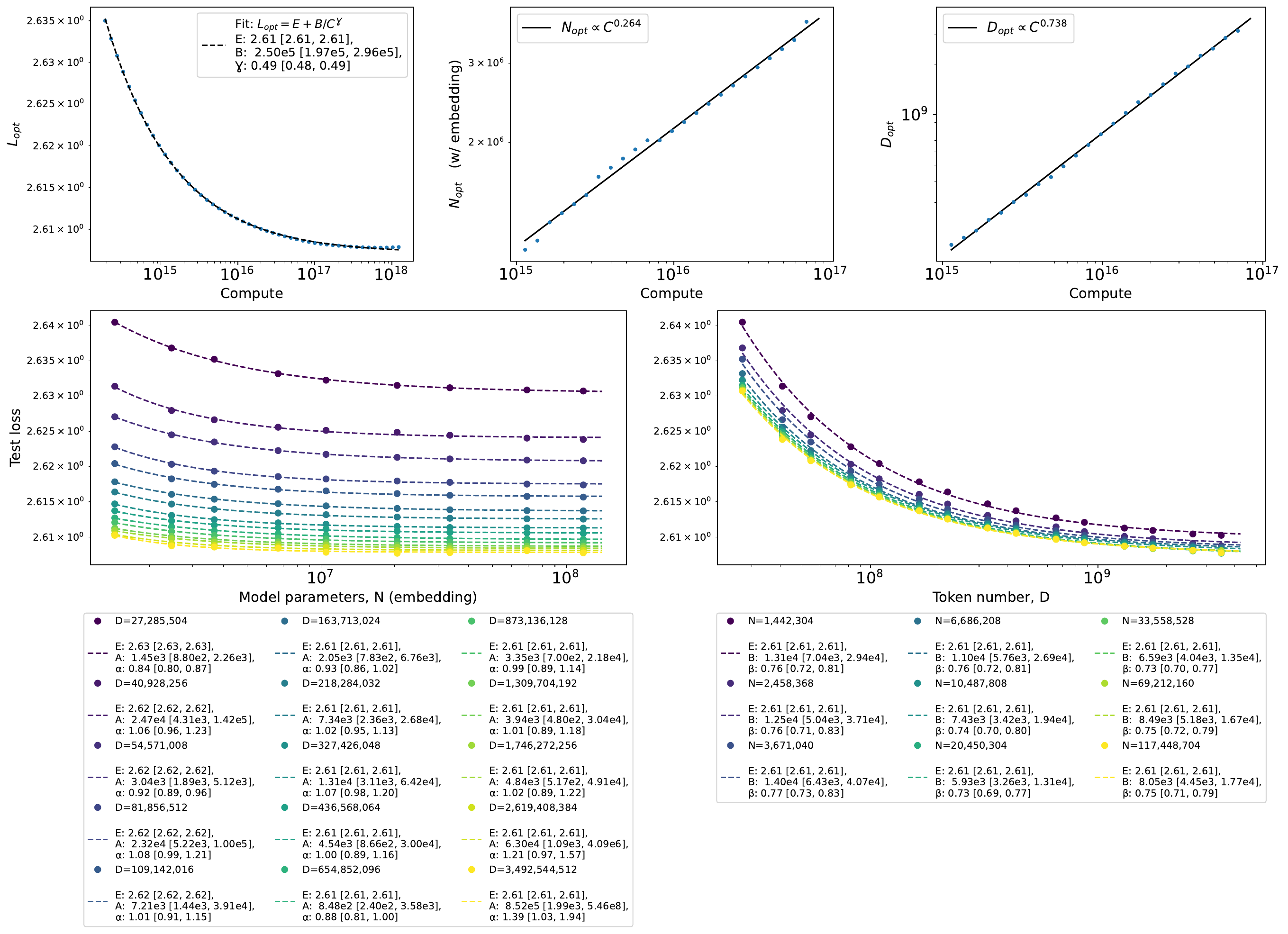}
    \caption{Scaling laws for 2-layer transformers trained on next token prediction on unbiased random walks on an Erdös-Renyi graph, with $8$K nodes and $50$K edges. Neither the random walks nor the graph exhibits any power laws. Data in bottom two plots are fit to Eq. \ref{1dlossfit}. Mean $\overline{\alpha_D} = 1.028$ with standard deviation $0.129$. Mean $\overline{\beta_N} = 0.749$ with standard deviation $0.014$. Average MSE for $L(N)_D$ 1d power law fits is $1.07\times 10^{-8}$, compared to $7.03 \times 10^{-8}$ for best exponential fit. Average MSE for $L(D)_N$ 1d power law fits is $8.86 \times 10^{-8}$, compared to $2.13 \times 10^{-6}$ for best exponential fit. Brackets indicate 95\% confidence intervals obtained from bias-corrected and accelerated bootstrap method (see Appendix \ref{app:fitting} for details). 
    }
    \label{fig:ER8k}
\end{figure}

\subsection{Our contributions}

\begin{figure}[t]
  \centering
  \includegraphics[width=\linewidth]{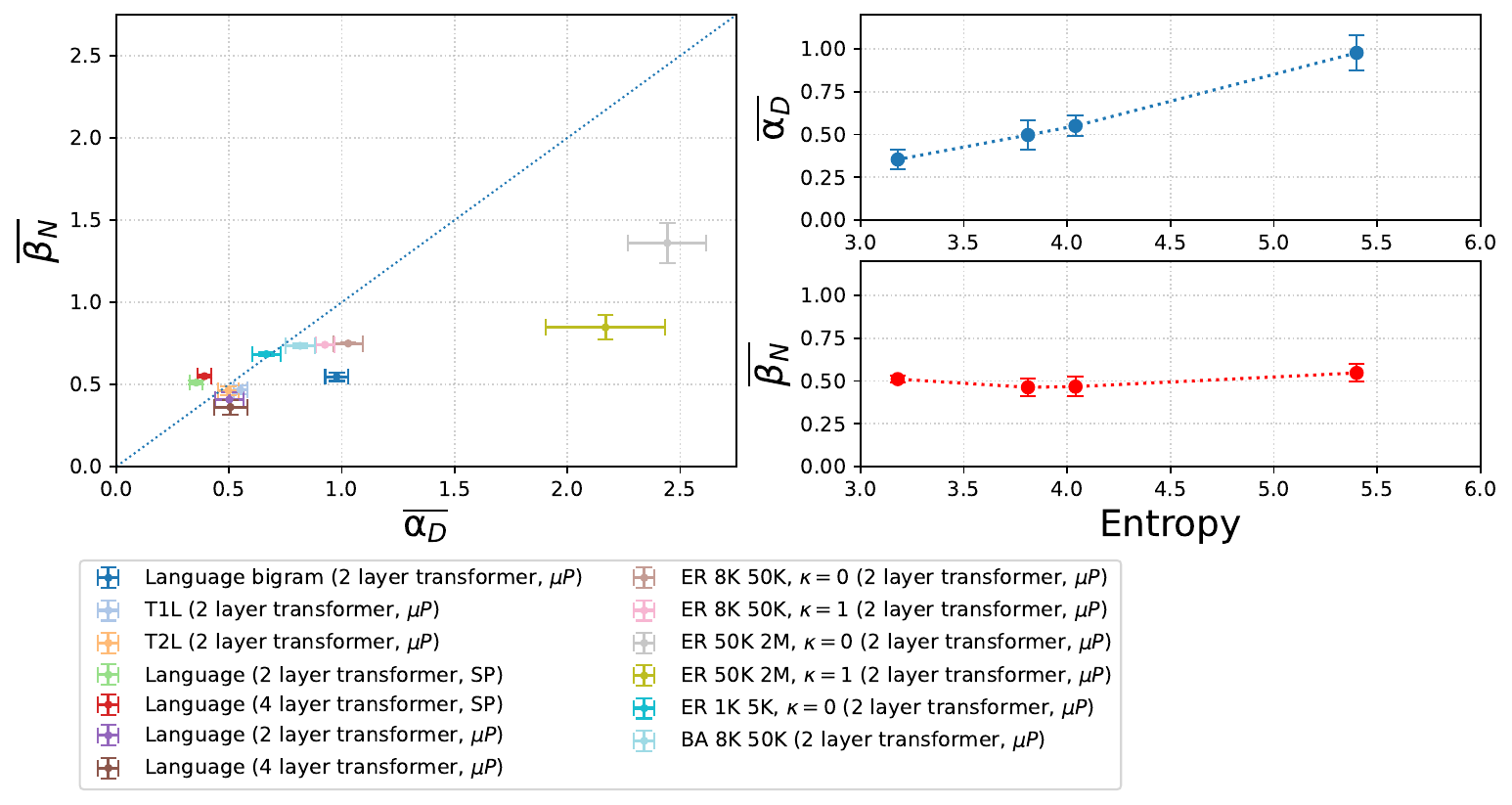}
  \caption{\label{fig:dialing_exponents} {\bf Left}: Mean exponents $\overline{\alpha_D}$ and $\overline{\beta_N}$ for all experiments reported in this paper. Legend is in the format <dataset> (model). Language refers to Fineweb-edu. $\overline{\alpha_D}$ and $\overline{\beta_N}$ are averages of the best fit exponents $\alpha_D$ and $\beta_N$ over $D$ and $N$. Error bars indicate standard deviation of the best fits $\alpha_D$, $\beta_N$ across different $N$ and $D$ respectively. {\bf Right}: $\overline{\alpha_D}$ and $\overline{\beta_N}$ for 2-layer transformer experiments on language, T1L, T2L, and language bigrams, demonstrating monotonic evolution of $\overline{\alpha_D}$ with approximate entropy of the dataset, along with relative stability of $\overline{\beta_N} \approx 0.5$. 
  }
\end{figure}

Our main contributions are as follows. 
\begin{itemize}
\item We demonstrate that transformers trained to perform next-token prediction on random walks from random graphs, such as Erdös-Renyi and Barabási-Albert graphs, exhibit neural scaling laws. In particular, we empirically demonstrate the first example of scaling laws even when the input data has no power law structures at all (Fig. \ref{fig:ER8k}).

\item We systematically reduce the complexity of the language setting by training on progressively simpler generative models of language, measuring the neural scaling laws at each level of complexity (Fig. \ref{fig:dialing_exponents}).  

\item We critically revisit the original neural scaling law results for LLMs and (i) argue that the best fits are one-dimensional fits of the form shown in Eq. \ref{1dlossfit}, in contrast to certain two-dimensional parametric fits that are common in the literature, and (ii) demonstrate a simple neural network regression method to obtaining compute optimal scaling laws, which is different from methods presented in the literature to date, and which gives higher accuracy for fitting loss curves (Figs. \ref{fig:chinchilla_mse_bar},\ref{fig:2layer_language_muP_mse_bar},\ref{fig:4layer_language_muP_mse_bar}).  

\item We demonstrate that the scaling laws for language found in prior work can be accounted for entirely within a 2-layer transformer with a context length of 100. We show in our experiments that the important discrepancy between the Kaplan and Chinchilla scaling laws can be  recovered in terms of whether embedding parameters are included \citep{porian2024resolving,pearce2024reconciling}. We also show preliminary results that maximal update parameterization \citep{yang2021tensor} might be more parameter-efficient than standard parameterization (Figs. \ref{fig:2layer_muP_language},\ref{fig:4layer_muP_language}), which would imply that compute-optimal scaling with $\mu P$ is not at constant number of tokens per parameter.  
\end{itemize}

\section{Neural scaling laws}

The most well-known and well-characterized neural scaling laws concern the test loss $L(N,D)$, where $N$ and $D$ denote the number of model parameters and the dataset size, respectively. In the context of LLMs, $D$ denotes the number of tokens, and $N$ sometimes excludes the embedding parameters. $L(N,D)$ is taken to be the optimal loss over a hyperparameter sweep:
\begin{align}
    L(N,D) = \min_{\{\xi\}} L(N,D;\{\xi\}), 
\end{align}
where $\{\xi \}$ denote optimization hyperparameters such as learning rate, batch size, weight decay, etc. In practice, one minimizes over a finite grid of hyperparameters. Furthermore, various architectural details, such as aspect ratio, number of attention heads, and so on, are sometimes ignored or held fixed, depending on the study. We denote $L(N)_D$ to be the test loss as a function of $N$ with $D$ held fixed and analogously for $L(D)_N$. Empirically, the test loss is well fit over several orders of magnitude by the equations
\begin{align}
\label{1dlossfit}
    L(N)_D = E_D + A_D N^{-\alpha_D}, \;\;\; L(D)_N = E_N + B_N D^{-\beta_N}
\end{align}
for sufficiently large $N$ or $D$, respectively. Note that all parameters in this fit depend on the variable that is held fixed. In some cases in the literature, the offsets $E_D$ and $E_N$ are not included, which leads to significant underestimation of the exponents $\alpha_D$ and $\beta_N$ (see Fig. \ref{fig:pythia_chinchilla}). 

It is common to consider two-dimensional parametric fits of the form \citep{hoffmann2022training,bhagia2025establishingtaskscalinglaws, muennighoff2023scaling,gadre2024overtraining,zhang2024map,kang2025demystifyingsyntheticdatallm}
\begin{equation}
\label{2dChinchillaEq}
    L(N,D) = E + \frac{A}{N^\alpha} + \frac{B}{D^\beta},
\end{equation}
which we refer to as the 2d Chinchilla formula. As we will see, while Eq. \ref{2dChinchillaEq} is useful for recovering compute optimal scaling curves and has the advantage of being interpretable, it often yields significantly worse fits than other methods, lacks theoretical support and -- in some cases -- is at odds with theoretical results in simple models \citep{paquette2024fourplus3}. \citep{kaplan2020scaling} considered a two-dimensional fit of the form 
\begin{equation}
\label{kaplan_eq}
L(N,D) = \left[\left(\frac{N_c}{N}\right)^{\alpha/\beta} + \frac{D_c}{D} \right]^\beta 
\end{equation}
for the case of early stopping, although this formula has not been adopted in the literature. The motivation for this formula was to have an analytic expansion in $1/D$ in the large $D$ limit, although there are situations \citep{hestness2017deep,hutter2021learningcurvetheory} where one expects fractional powers of $D$ in the large $D$ limit. A large variety of other low-dimensional parametric formulas have been considered; see Table 2 of \citep{li2024mis} for a catalog.  

The compute in FLOPs (floating point operations) can be expressed exactly in terms of $N$, $D$, and number of training steps \citep{kaplan2020scaling}. It approximately follows the formula $C(N,D) \approx 6 N B S$, where $S$ is the number of steps and $B$ is the batch size. For the standard case of 1 epoch training, which we consider hereafter, $D = BS$ and $C(N,D) \approx 6 N D$. The compute optimal loss is given by minimizing the loss over $N$ and $D$ subject to a fixed compute:
\begin{align}
\label{compute_optimal_1}
    L_{\text{opt}}(C) = \min_{N,D | \text{fixed } C} L(N,D) .
\end{align}
The optimal $N$ and $D$ for a fixed compute are then given by
\begin{align}
\label{compute_optimal_2}
    N_{\text{opt}} = \text{argmin}_N L(N, D(C,N)), \;\; D_{opt} = \text{argmin}_D L(N(D,C), N),
\end{align}
where in the equation above $N(D,C) \approx C/6D$ refers to $N$ as viewed as a function of $D$ and $C$, and similarly for $D(N,C) \approx C/6N$. $L_{\text{opt}}$, $N_{\text{opt}}$, and $D_{\text{opt}}$ are empirically fit well to
\begin{align}
\label{compute_optimal_3}
    L_{\text{opt}} = E_C + K C^{-\gamma}, \;\;\;\;
    N_{\text{opt}} = N_0 C^a, \;\;\;\; D_{\text{opt}} = D_0 C^b ,
\end{align}
with $a + b \approx 1$. 

\subsection{Methodology}
\label{sec:methodology}

To extract the scaling laws, we first fit our results for the loss to the 1d parametric fits, Eq. \ref{1dlossfit}. For details regarding the fitting procedure, see Appendix \ref{app:fitting}. We note that fitting a power law does not in general correspond to fitting the loss to a line on a log-log plot. It is crucial to include the irreducible loss, without which one can easily underestimate scaling exponents by over $2\times$ in practice (see Fig. \ref{fig:pythia_chinchilla}). 

An important discussion is to what extent the power law fits are accurate \citep{clauset2009}, especially for their stated purpose of extrapolating predictions for the test loss out to several orders of magnitude in $N$, $D$, or $C$ beyond which there is data for $L(N,D)$.\footnote{Note that the number of orders of magnitude covered by $N$ can change significantly depending on whether $N$ includes embedding parameters. Compare for example Fig. \ref{fig:language_sp_embed} with \ref{fig:language_sp_nonembed}. } In this paper, to validate the power law fit, we compare it to an exponential fit of the form $f(x) = a +  b e^{-cx}$. The ratio between the mean squared error (MSE) of the power law and exponential fits are depicted in Fig. \ref{fig:mse_ratios}. In all cases studied in this paper, the best power law fits for $L(D)_N$ have factors of 50-100x better MSEs than the best exponential fits; for $L(N)_D$, with one exception the power law fits have factors of 5-100x better MSEs than the exponential fits. An important message of our paper is that next-token prediction on various types of random walks gives power law fits to the loss that are nearly as good from the perspective of MSE -- and in some cases better -- than those for natural language.

To obtain compute optimal fits requires more analysis, because in practice we have not sampled enough values of $N$, $D$. In the next step, we obtain a prediction $\hat{L}(N,D)$ for values of $N$,$D$ that are not in the raw dataset. To obtain this prediction, instead of using the 2d fits of Eq. \ref{2dChinchillaEq}, \ref{kaplan_eq} as is common in the literature, we use an alternative technique -- regression with a 3-layer fully connected neural network (for details see Appendix \ref{app:fitting}). We demonstrate that this method (along with a kernel regression method) gives significantly better train/validation errors than the 2d Chinchilla fit that is widely used in the literature (see Figs. \ref{fig:chinchilla_mse_bar},\ref{fig:2layer_language_muP_mse_bar},\ref{fig:4layer_language_muP_mse_bar}). Since the 2d Chinchilla fit does not have a strong theoretical basis, we suggest eschewing it in favor of more standard ML regression techniques. Given $\hat{L}(N,D)$, we can then use a sufficiently fine grid for $N,D$ to extract compute optimal scaling laws following Eqs. \ref{compute_optimal_1}, \ref{compute_optimal_2}, \ref{compute_optimal_3}. 

In all of our figure captions we report the mean exponents $\overline{\alpha_D}$ and $\overline{\beta_N}$, which are the average of the best-fit exponents $\alpha_D$ and $\beta_N$ across different values of $D$ and $N$. We also report the standard deviation for the best fit $\alpha_D$ and $\beta_N$ across different values of $D$ and $N$, which are the error bars in Fig. \ref{fig:dialing_exponents}. Note that these standard deviations do not take into account the confidence intervals for each fit, which are shown in brackets in the figure legends.  

Details of our experiments are presented in Appendix \ref{app:exp_details}. We use a decoder-only GPT-style transformer to perform next-token prediction. We use settings characteristic of standard language pretraining \citep{brown2020language,hoffmann2022training}, including 1 epoch training with a learning rate schedule consisting of linear warmup and cosine decay. In all of our figures, ``Compute" refers to the number of FLOPs (floating point operations), ``$N$ (embedding)" refers to the total number of parameters including embedding parameters, while ``$N$ (non-embedding)" excludes the embedding parameter count. 

\section{Random walks, $n$-grams, and random graphs}
\label{sec:ngrams}

The simplest non-trivial generative model of sequences is a Markov random walk on a graph. Consider a directed, weighted graph on $N$ nodes with weight matrix $W_{uv}$. We can sample a random walk by randomly starting at a node with some initial distribution and then iteratively sampling from the probability distribution $p(u|v) := W_{uv}/\sum_u W_{uv}$.

Observe that the above is equivalent to generating from a bigram model. Recall that an $n$-gram language model approximates the conditional joint distribution of text in terms of the $n-1$ most recent tokens: $p(x_t | x_{t-1}, \cdots, x_0) = p_n(x_t|x_{t-1}\cdots, x_{t-n+1})$. In the case of a bigram model ($n = 2$), this defines a directed, weighted graph, which we refer to as the bigram graph. For a vocabulary of size $V$, the bigram graph has $V$ nodes, and each directed edge $uv$ is associated with a weight $p_2(u,v)$. Sampling from a bigram model corresponds to sampling a random walk on the bigram graph, with transition probabilities $p(u|v) = p_2(u,v)/p_1(v)$, where $p_1(v) = \sum_{u} p_2(u,v)$ is the stationary distribution on nodes, and equivalently the unigram distribution of the sequence.  

This picture can be readily generalized to $n > 2$. In that case, the $n$-gram model defines a directed, weighted hypergraph with $V$ nodes. Directed $n$-hyperedges, which consist of ordered $n$-tuples $(x_1, \cdots, x_n)$, are associated with a weight $p_n(x_n, \cdots, x_1)$. Sampling from the $n$-gram model then consists of sampling a random walk on the hypergraph: given $n-1$ nodes $x_{t-1}, \cdots, x_{t-n+1}$, a hyperedge is chosen at random, giving the next node $x_t$ in the sequence. The initial $n-1$ nodes can be chosen from the stationary joint distribution on $n-1$ nodes.  

Next-token prediction on a random walk on an ordinary graph ($n =2$) is essentially a problem of learning a probability table. Given training sequences $\{x_1^{(i)}, \cdots, x_T^{(i)}\}$, for $i = 1, \cdots, N_{\text{seq}}$, the task requires the model to learn that they are controlled by bigram transition probabilities $p(u|v)$ and then to learn the probabilities themselves. The obvious algorithm to learn this is to form a prediction $\hat{p}$ by counting co-occurences in the training set, which is exactly how bigram models are learned \citep{JurafskyMartin2009}. If we consider a cross-entropy or MSE loss function $L$ comparing the prediction $\hat{p}$ to the ground truth $p$, then in expectation over a sample of size $D$ tokens,
\begin{align}
    \mathbb{E}[L] = E + B/D + O(1/D^2),
\end{align}
as shown in Appendix \ref{app:baseline}. For cross-entropy loss, $E$ is the entropy of the distribution $p$, while for MSE loss $E = 0$. $B$ is a constant depending on $|V|$. This provides a baseline against which we can compare our empirical results. Note that this can be thought of as the "variance-limited" regime for power law scaling of data in the language of \citep{bahri2021explaining}. 

An unbiased random walk on an undirected graph has the transition probability $p(u|v) = \frac{A_{uv}}{\text{deg}(v)}$, where $A$ is the adjacency matrix and $\text{deg}(v) = \sum_u A_{uv}$ is the degree of $v$. The stationary distribution is $p(u) = \frac{\text{deg}(u)}{2E}$, where $E$ is the number of edges in the graph. Recall that random walks have an exponentially decaying auto-correlation, with the time-scale for the exponential decay set by the spectral gap of the transition matrix. 

\subsection{Random graphs: Erdös-Renyi and Barabási-Albert ensembles}

\begin{figure}[t]
  \centering
  \includegraphics[width=\linewidth]{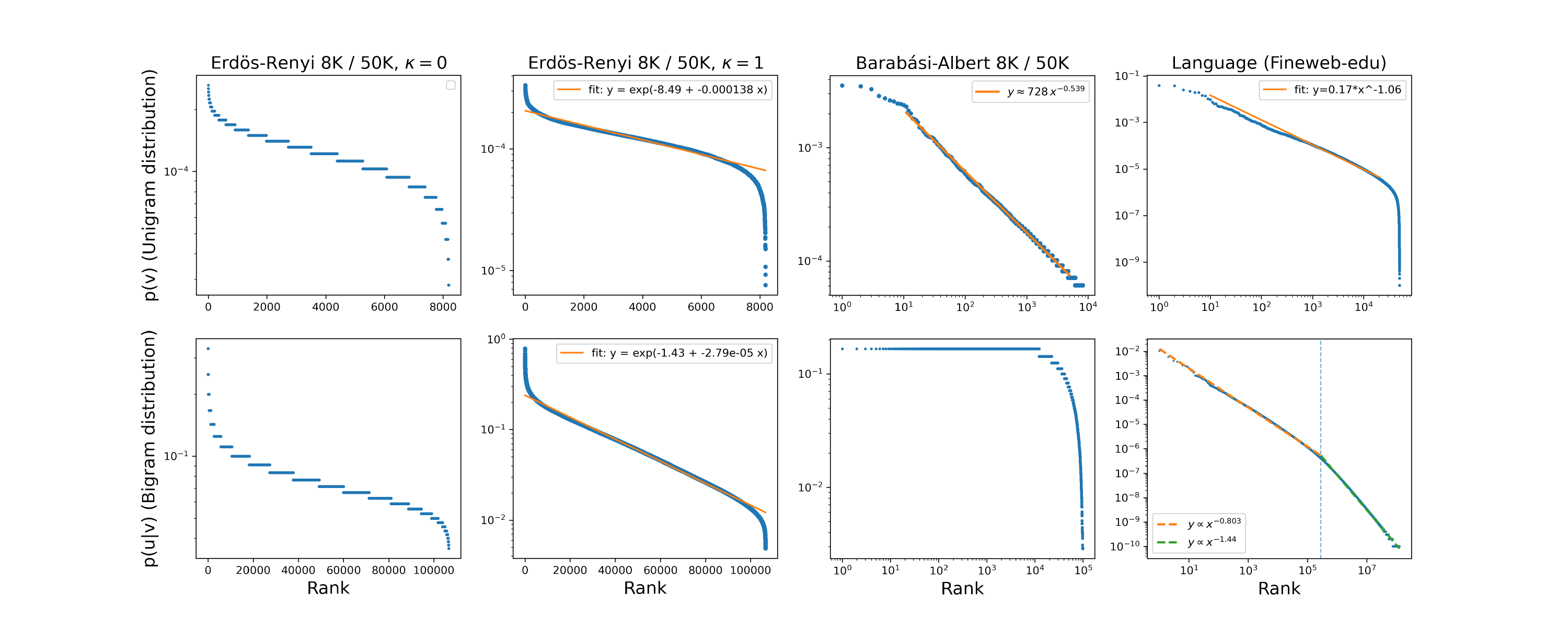}
  \caption{Stationary distribution on nodes $p(v)$ (unigram distribution) and transition probability distributions $p(u|v)$ (bigram distribution) plotted against the rank for various representative cases studied in this paper. Note that $p(v)$ is equal (up to an overall constant) to the degree distribution of the graph when $\kappa = 0$. Left two columns are plotted on log-linear scale, and the right two columns are on log-log scale. ER $\kappa = 0$ case is tightly concentrated around its mean probability, as indicated by roughly constant size plateaus. BA ($\kappa = 0)$) has power laws in both unigram and bigram distributions; the unigram exponent is approximately equal to the theoretical expectation $1/(\gamma - 1) = 0.5$. The power law in the bigram is more clear if plotted in reverse rank order (not shown), but can be observed from the decreasing width of the plateaus. Language results are from Fineweb-edu using GPT-2 tokenizer; unigram distribution is a standard Zipf plot, while bigram distribution shows broken power law. 
}
  \label{fig:graph_plots}
\end{figure}

In order to understand better the scaling laws of language models, it is interesting to compare their performance to a simple baseline of learning random walks on random graphs. 

\subsubsection{Erdös-Renyi graphs}

The simplest class of random graphs are Erdös-Renyi graphs. The number of nodes is denoted $N$. The adjacency matrix of the graph is chosen randomly such that $A_{ij} = 1$ with probability $p$ and $0$ with probability $1-p$. The degree distribution of the nodes is tightly concentrated around the expected degree, $\langle k \rangle = p(N-1)$, and the expected number of edges is $\langle E \rangle = \langle k \rangle N/2 = pN(N-1)/2$. The spectrum of the adjacency matrix obeys a semicircle law, which is typical of random symmetric matrices. Neither the degree distribution nor the spectrum of the graph has any power laws. In Fig. \ref{fig:graph_plots} we present the degree distribution (equivalent to the unigram distribution for unbiased walks) and the bigram distribution ($p(u|v)$) for an unbiased random walk on an ER graph with $N = 8,192$ and $E = 53,292$ that we use in our experiments, demonstrating the absence of any power law structure in the input data correlations.  

Our first main result is presented in Fig. \ref{fig:ER8k}, demonstrating the existence of neural scaling law behavior. This result demonstrates that neural scaling laws are a generic feature of learning sequence models with cross-entropy loss, \it independent of whether the data itself has any power law structure. \rm These models were trained with maximal update parameterization ($\mu P$) with context length of $50$; additional experimental details are presented in Appendix \ref{app:exp_details}.  

We have studied unbiased random walks on ER graphs of varying sizes. Fig. \ref{fig:er50k2M_alpha0} demonstrates results from an ER graph with $N = 50,000$ and $E = 2$M edges. We also present results for ER graphs with $1$K nodes and $5$K edges in Fig \ref{fig:ER1k_kappa0}; this in particular demonstrates that scaling laws in these settings can already be observed at relatively small scales, which should be accessible at relatively low compute levels ($10^{13} - 10^{16}$ FLOPs in our experiments). 

\begin{figure}[t]
  \centering
  \includegraphics[width=\linewidth]{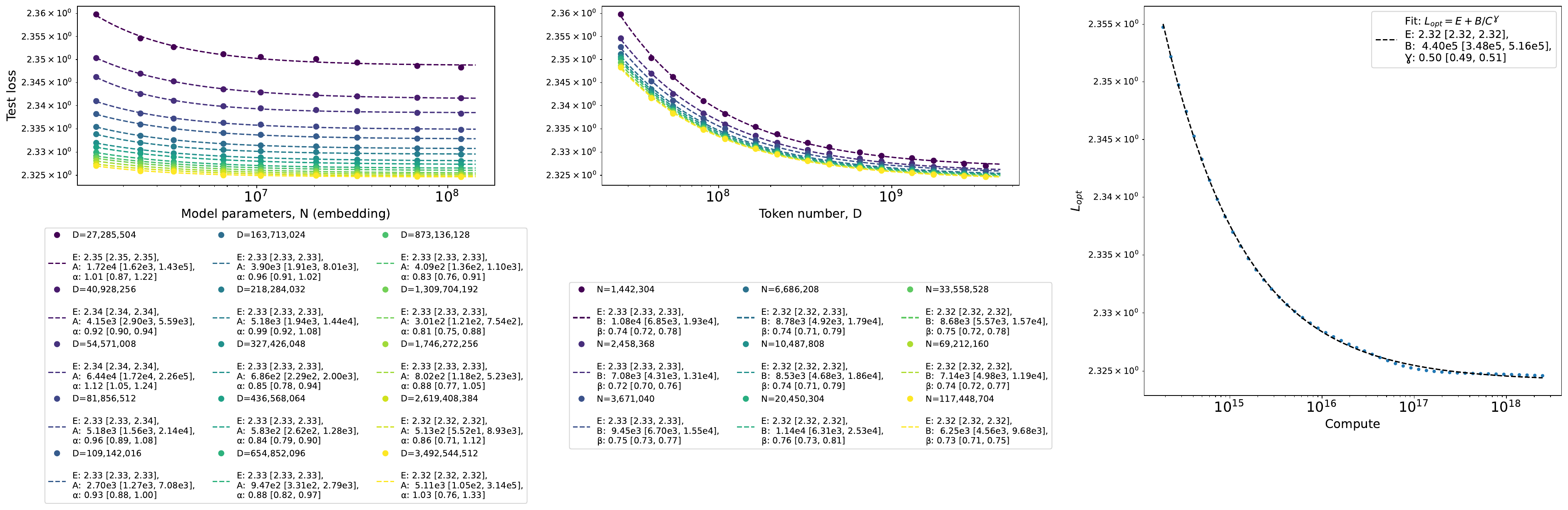}
  \caption{Scaling laws for Erdos-Renyi graph with 8K  nodes and 50K edges, with power-law transition probabilities set by $\kappa = 1$. Mean exponent $\overline{\alpha_D} = 0.925$ with standard deviation $0.084$. Mean exponent $\overline{\beta_D} = 0.741$, with standard deviation $0.010$. Average MSE for $L(N)_D$ 1d power law fits is $1.76 \times 10^{-8}$, compared to $1.03 \times 10^{-7}$ for best exponential fit. Average MSE for $L(D)_N$ 1d power law fits: $5.15 \times 10^{-8}$, compared to $2.12 \times 10^{-6}$ for best exponential fit. 
  }
  \label{fig:ER8k_kappa1}
\end{figure}

\subsubsection{Biased random walks: dialing into power laws}

We can controllably tune the stationary distribution (equivalently, unigram distribution) $p(v)$ and the transition probabilities $p(u|v)$ (equivalently, bigram distribution) by considering biased random walks. We consider sampling a weight matrix $W_{uv}$ from a power-law distribution $\text{Pr}(W_{uv} = k) \propto k^{-\kappa}$, for $k \in [k_{\text{min}}, k_{\text{max}}]$ and $0$ otherwise (when the edge $uv$ is present in the graph). For $\kappa > 1$, the transition probabilities follow a power law distribution, $P(u|v) \propto (1/r_{(u|v)})^{\frac{1}{\kappa - 1}}$, where $r_{(u|v)}$ is the rank of the bigram $(u|v)$, sorted from highest to lowest in probability. $\kappa = 1$ is the marginal case, where the probabilities and their ranks follow an exponential (log-linear) relationship: $P(u|v) = P_0 e^{- s r_{(u|v)}}$, for constants $s, P_0$. 

We present results for the marginal case $\kappa = 1$. The unigram and bigram distributions for the ER graph with 8K nodes and 50K edges are shown in Fig \ref{fig:graph_plots}, which shows that $\kappa = 1$ smooths the unigram and bigram distributions compared to the case $\kappa = 0$. This is an interesting intermediate case: on the one hand the data correlations still lack power laws, on the other hand they sit in between the $\kappa = 0$ case and the case where the unigram and bigram distributions do have power laws. 

The scaling law results for the marginal case $\kappa = 1$ are shown in Fig. \ref{fig:ER8k_kappa1} for the 8K / 50K graph and in Fig. \ref{fig:er50k2M_alpha1} for the 50K / 2M graph. The primary observation is that the quality of the power law fits are improved (see, e.g. Fig. \ref{MSE_ratios}), suggesting that the structure of the data correlations 
can help crystallize the power law scaling in the loss. It would be interesting to continuously tune the exponent $\kappa$ and see the evolution in the scaling exponents and the quality of the fit, which we leave for future work.

\begin{figure}[t]
  \centering
  \includegraphics[width=\linewidth]{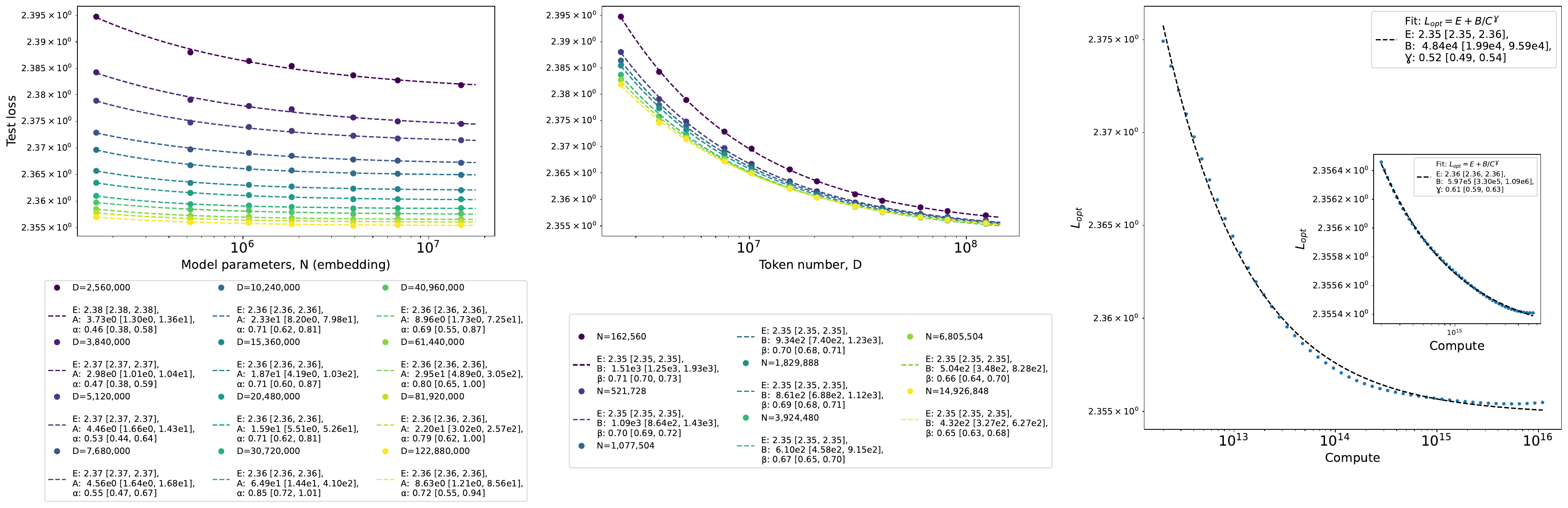}
  \caption{Scaling laws for unbiased ($\kappa = 0$) random walk on Erdos-Renyi graph with 1K  nodes and 5K edges. Mean exponent $\overline{\alpha_D} = 0.665$ with standard deviation $0.125$. Mean exponent $\overline{\beta_D} = 0.684$, with standard deviation $0.021$. Average MSE for $L(N)_D$ 1d power law fits is 
  $2.41e-08$, compared to $1.96 \times 10^{-7}$ for best exponential fit. Average MSE for $L(D)_N$ 1d power law fits: $3.37 \times 10^{-8}$, compared to $1.97 \times 10^{-6}$ for best exponential fit. }
  \label{fig:ER1k_kappa0}
\end{figure}

\subsubsection{Scale-free graphs: Barabási-Albert ensemble}

Another way to introduce power law structures in the data is to introduce it in the connectivity structure of the graph itself. A scale-free graph \citep{barabasi1999emergence} is defined such that a randomly chosen node has degree $k$ with probability $\propto k^{-\gamma}$ for large $k$, with $\gamma > 1$. Typical examples of approximately scale-free graphs include the world wide web inward-directed edges, protein-protein interactomes, and the in-degree of citation networks for academic papers. Co-occurrence matrices from natural language corpora in particular have been shown to determine a scale-free, small-world graph \citep{cancho2001small}. 
One way to construct a scale-free graph is via the Barabási-Albert preferential attachment model \citep{barabasi1999emergence}, which gives $\gamma = 3$. We use this model to construct a scale-free graph with $8,192$ nodes and $49,131$ edges. We display the power-law unigram and bigram distributions for the unbiased ($\kappa = 0$) walk in Fig. \ref{fig:graph_plots}. The resulting scaling laws for learning unbiased random walks on this graph are presented in Fig. \ref{fig:ba}. Qualitatively the results are similar to the ER graphs without power law structures, and in fact the power law fit is somewhat worse for $L(N)_D$ (Fig. \ref{fig:mse_ratios}).

\section{Dialing up complexity: from random graphs to natural language}

It is of great interest to understand the origin of the neural scaling laws and to what extent the scaling exponents can be altered by the architecture, data and learning algorithms. One way forward is to develop  methods to define and tune the complexity of the dataset and to study how the scaling exponents vary. At the high level of complexity are real-world datasets, like natural language, math, and coding datasets. At the low end of complexity are bigram models, which are equivalent to Markov random walks on graphs. In this paper we consider a monotonic dialing up of the complexity of language models. At the lowest level of complexity we start with a bigram language model. At an intermediate level of complexity, we consider a synthetic dataset generated from sampling the outputs of an $n$-layer transformer trained on FineWeb-edu \citep{penedo2024fineweb}. For a sufficiently large transformer, given the stunning success of transformer-based large language models, these sequences should approximate natural language quite closely. For small transformers they presumably provide a synthetic dataset of intermediate complexity. In this paper we study the case $n = 1,2,4$, and refer to these datasets as Transformer-generated $n$-layer (TnL) datasets. 

One can imagine other generative models that also interpolate complexity from bigram models to natural language: $n$-gram models for $n > 2$ and hierarchical graphical models of natural language. We leave the study of scaling laws for these cases to future work.

As a rough measure of the complexity, we use the entropy of the joint distribution of the sequences in the dataset. We think of this as a rough measure of complexity because a low complexity dataset has low signal in predicting $x_t$ from $x_{t-1}, x_{t-2},\cdots$, giving rise to many likely possibilities and therefore high entropy. On the other hand, a high complexity dataset has high signal in predicting $x_t$ from $x_{t-1}, x_{t-2},\cdots$, and therefore the entropy should be lowered. Indeed, it is easy to show that for an $n$-gram language model, the entropy monotonically decreases with $n$. 

While one can define and compute the entropy of an $n$-gram model, it is not feasible to compute the entropy of the sequences generated by TnL since it would require a number of samples that is exponential in the sequence length. Even worse, the entropy of natural language does not have an intrinsic definition, since there is no true generative model that one can use to define a joint distribution. We can, however, approximate the entropies by recalling that the cross-entropy loss is lower-bounded by the entropy of the reference distribution. Therefore, if we fit power laws $L(N)_D = E_D + A_D/N^{\alpha_D}$, and $L(D)_N = E_N + B_N/D^{\beta_N}$, the entropy of the reference distribution can be approximated as $E = \lim_{N \rightarrow \infty} E_N = \lim_{D \rightarrow \infty} E_D$. As a rough estimate, we take the smallest value of $E_N$ or $E_D$ that appear in our fits and use that as a proxy for the entropy $E$. This gives $E = 5.40, 4.04,3.81,3.18$ for the bigram langauge model, T1L, T2L, and natural language (Fineweb-edu), respectively, which is consistent with our expectation that these generative models have increasingly higher complexity. 

\subsection{Scaling laws for learning language bigrams}

\begin{figure}[t]
  \centering
  \includegraphics[width=\linewidth]{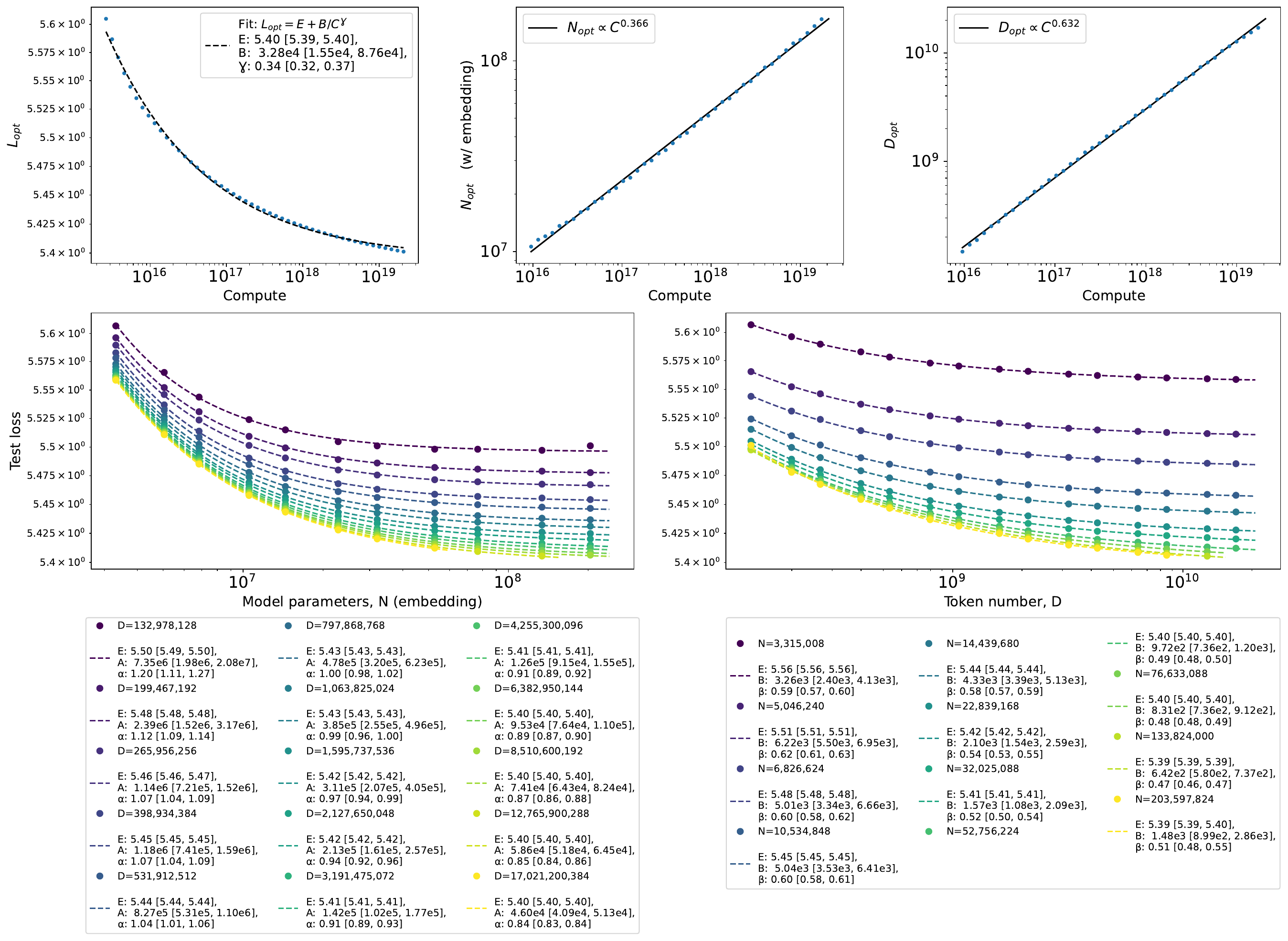}
  \caption{\label{2layermuPlanguageLD_embed}2 layer transformer trained on language bigram with $\mu P$. The $\alpha_D$ have a mean of $\overline{\alpha_D} = 0.977$ and a standard deviation of $0.102$. The $\beta_N$ have a mean of $\overline{\beta_N}  = 0.547$ and a standard deviation of $0.052$. Average MSE for $L(N)_D$ 1d power law fit is $5.22 \times 10^{-7}$, as compared to $2.91 \times 10^{-5}$ for the best exponential fit. Average MSE for $L(D)_N$ 1d power law fits is $1.66 \times 10^{-7}$, as compared to $6.51 \times 10^{-5}$ for the best exponential fit. 
 }
  \label{fig:bigram_scaling_laws}
\end{figure}

Here we discuss our results for scaling laws for next-token prediction on sequences generated from a bigram language model. To define our bigram langauge model, we use the GPT-2 tokenizer \citep{radford2019language}, which has a vocabulary of size $50,257$. We train our bigram model on the Fineweb-edu-10B dataset \citep{penedo2024fineweb}. This dataset has approximately $130$M unique bigrams. Fig. \ref{fig:graph_plots} (bottom right) displays the bigram probabilities $p(u,v)$ vs. their rank, showing a (broken) power-law distribution. As depicted in Fig. \ref{fig:bigram_stats}, a large number of bigrams appear only a handful of times; for our work to be computationally tractable, we remove all bigrams that appear $5$ times or fewer, leaving around $33$M unique bigrams. Therefore, our truncated bigram model corresponds to a graph with approximately $50$K nodes and $33$M edges, and with a (broken) power law weight matrix. 

The scaling laws for learning using a 2-layer transformer with maximal update parameterization ($\mu P$), a context length of $50$, and a fixed batch size of $100$ sequences are shown in Fig. \ref{fig:bigram_scaling_laws}. The exponents $\alpha_D$ lie in the range $0.84 - 1.2$. The mean and standard deviation of the best fit exponents across the different values of $D$ are $\overline{\alpha_D} = 0.977$, $\sigma_{\alpha_D} = 0.102$ respectively. The $\beta_N$ lie in the range $0.47 - 0.60$, with mean $\overline{\beta_N} = 0.547$ and standard deviation $\sigma_{\beta_N} = 0.052$. While the $\alpha_N$'s are significantly higher than natural langauge, the $\beta$'s are much closer. 
Interestingly, this language bigram example shows the clearest power law fits. As depicted in Fig. \ref{fig:mse_ratios}, the ratio of the MSE between the best power-law and best exponential fits are $0.0179$ and $0.0026$, respectively. 

The compute optimal curves also demonstrate that learning the bigram model is much more compute efficient than natural language, as the exponent $\gamma$ is substantially larger, and also more parameter efficient, as $a$ is almost half as large as $b$.

\subsection{Scaling laws for learning TnL-Language}

\begin{figure}[t]
  \centering
  \includegraphics[width=\linewidth]{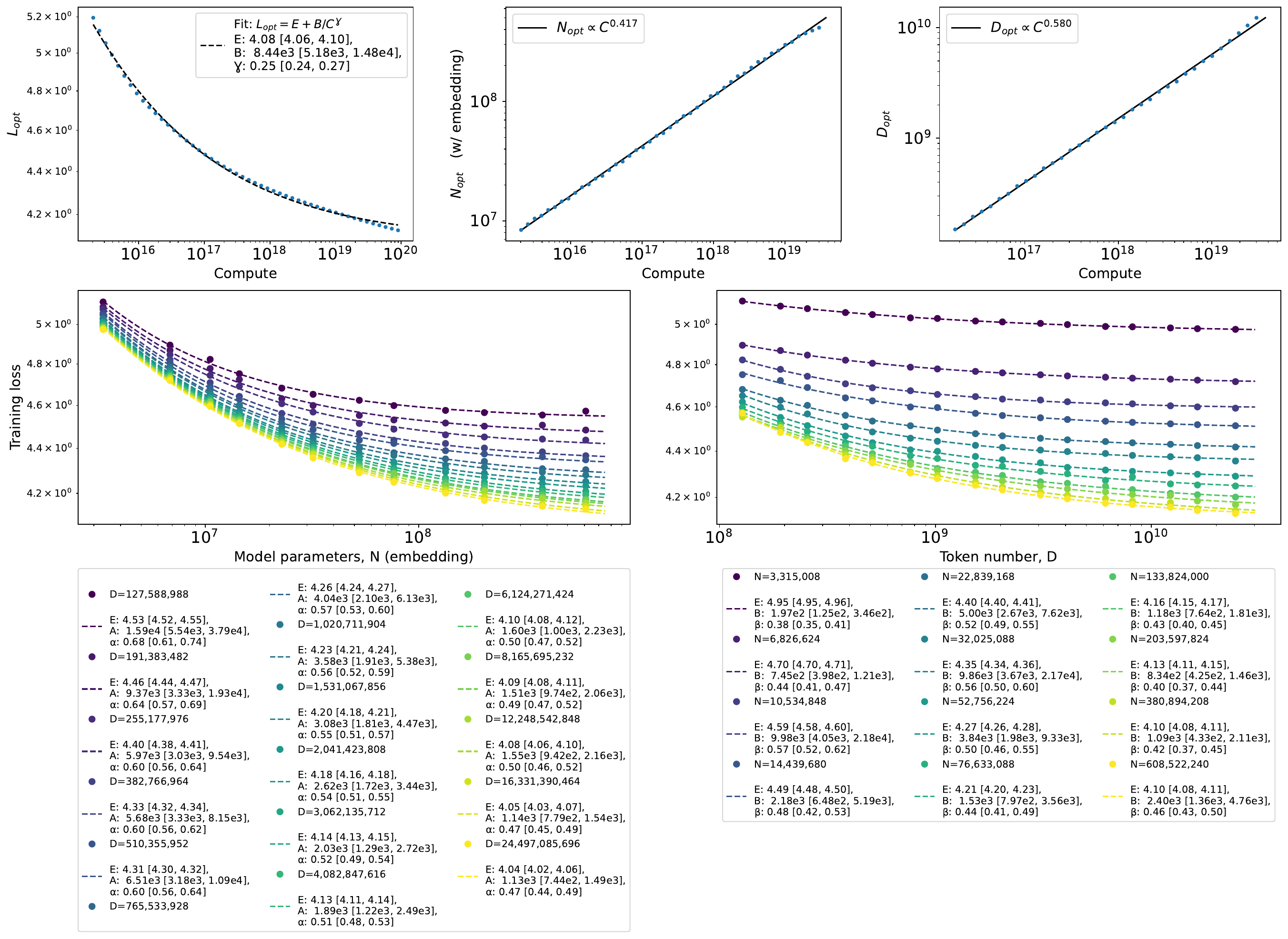}
  \caption{\label{2layermuPlanguageT1L} Scaling laws for 2 layer transformers learning T1L. The $\alpha_D$ have a mean of $0.550$, with a standard deviation of $0.059$. The $\beta_N$ have a mean of $0.467$, with standard deviation $0.058$. The average MSE for the $L(N)_D$ 1d power law fits is $6.84 \times 10^{-5}$, while the average MSE for the best fit exponential fit is $1.49 \times 10^{-3}$. Average MSE for $L(D)_N$ 1d power law fits is $2.13 \times 10^{-5}$, compared to $1.27 \times 10^{-3}$ for the best exponential fit.
  }
  \label{fig:T1L}
\end{figure}

\begin{figure}[t]
  \centering
  \includegraphics[width=\linewidth]{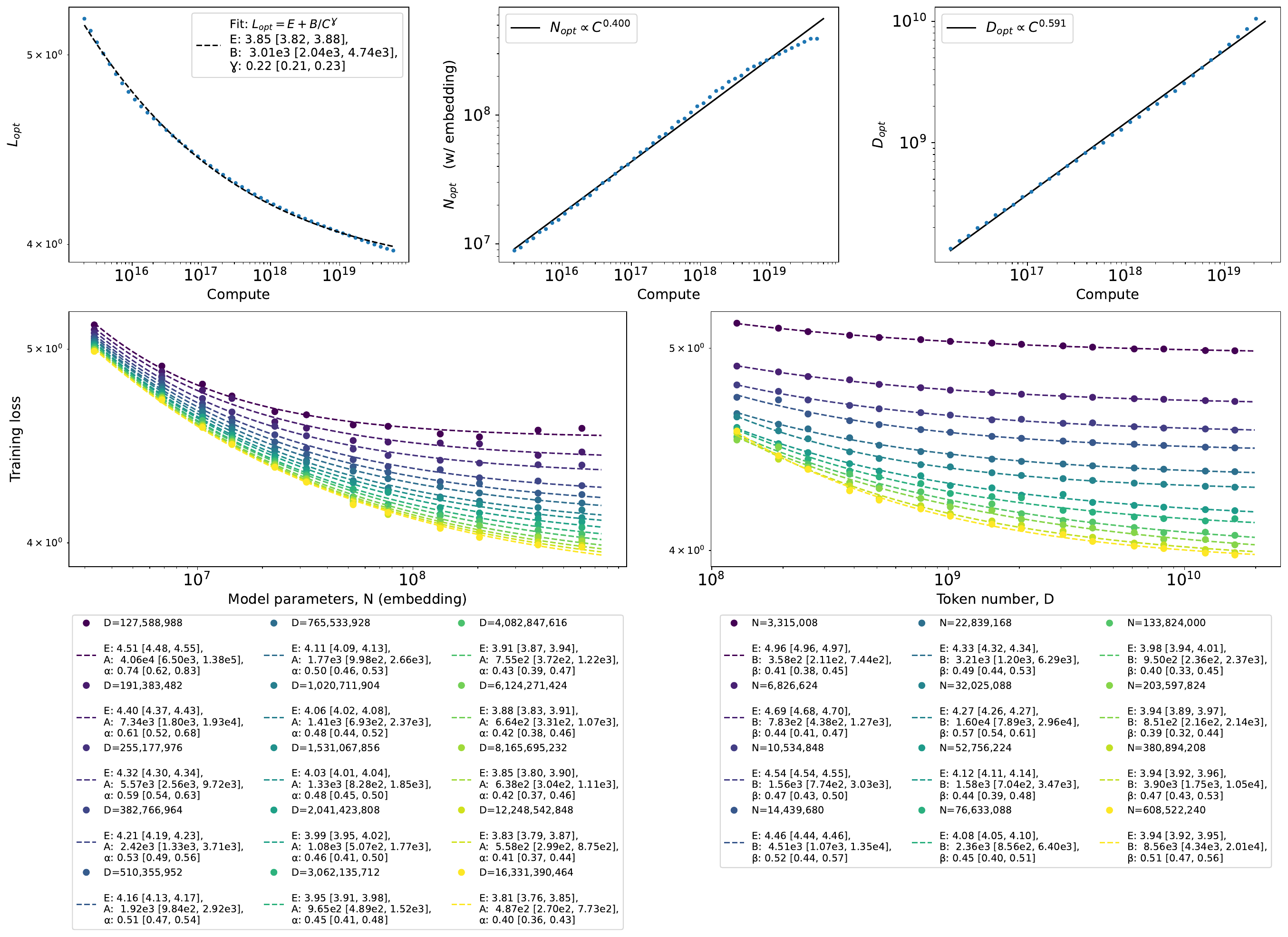}
  \caption{\label{2layermuPlanguageT2L} 2 layer transformer train on T2L. The mean $\alpha_D$ is $0.497$, with standard deviation $0.088$. The mean $\beta_N$ is $0.463$ with standard deviation $0.052$. The average MSE for $L(N)_D$ 1d power law fits is $1.91 \times 10^{-4}$, compared to $2.03\times 10^{-3}$ for the best exponential fit. Average MSE for $L(D)_N$ 1d power law fits is $6.22 \times 10^{-5}$, compared to $1.52 \times 10^{-3}$ for the best exponential fits. 
  }
  \label{fig:T2L}
\end{figure}

Here we discuss our results for scaling laws for 2-layer transformers learning TnL sequences. These experiments had the same settings as the language bigram experiments: $\mu P$ parameterization with a context length of $50$ and a fixed batch size of 100 sequences.  Fig. \ref{fig:T1L} and \ref{fig:T2L} show the case with $n = 1$ and $2$ respectively. 

As summarized in Fig. \ref{fig:dialing_exponents}, we found that the mean of exponents $\overline{\alpha_D}$ interpolate between the language bigram and natural language results. 

The results for T4L were nearly identical to those for T2L, so we omit them from the presentation. We expect the reason is that a 2-layer transformer may not be able to clearly distinguish between sequences generated from a 2-layer and 4-layer transformer

\section{Revisiting scaling laws for natural language modeling}

In this section we present our results on neural scaling laws for LLM pretraining. Scaling laws for natural language modeling have been well-studied in the literature; the primary purpose of our discussion here is twofold. One is to to explain our methodology while providing a critical analysis of some widely used methodology in the literature. The other is to demonstrate how the results in the previous sections directly compare to the setting with natural language. Specifically, we present four main points:
\begin{enumerate}
    \item Dropping the constants $E$ (referred to as the irreducible loss) in the 1d fits, as done in some prior works, leads to dramatic underestimation of the scaling exponents and significantly poorer fits. Consequently, power-law scaling in the loss should not show up as a clean line on a log-log plot, but rather have curvature, as shown in all our results. 
    \item The two-dimensional Chinchilla formula in general gives poor fits compared to other standard ML methods, in addition to having no theoretical basis. We suggest replacing the 2d Chinchilla formula with neural network regression to fit $L(N,D)$, which is the methodology used in this paper to obtain compute-optimal scaling results.
    \item Many prior results in neural scaling laws -- including the discrepancy between \citep{hoffmann2022training} and \citep{kaplan2020scaling} -- can be largely recovered from training shallow transformers (2 layers) with short context lengths (context length 100); in our experiments, resolving the discrepancy does not require changing the learning rate schedule as discussed in many works. 
    \item We find preliminary evidence that using maximal update $\mu P$ parameterization \citep{yang2021tensor} may be more parameter efficient for compute optimal training. 
\end{enumerate}

\subsection{Importance of the irreducible loss}

To make the first point (1) above, we consider six models in the Pythia sequence of LLMs \citep{biderman2023pythia}, which were trained on 300B tokens from the Pile dataset. Using the data provided by \citep{michaud2024quantizationmodelneuralscaling}, we consider the 1d fit $L(N)_D$ both with and without the offset $E_D$. Neglecting the offset $E_D$ as in \citep{michaud2024quantizationmodelneuralscaling} gives a nearly factor of two difference in the scaling exponent and almost an order of magnitude difference in the mean squared error (MSE) of the fit. We note that an exponent of $\alpha = 0.07$ was presented in \citep{henighan2020scaling} (see top left of Fig. 3 in \citep{henighan2020scaling}) upon neglecting $E_D$, and we expect a similar conclusion to hold in that case as well. 

\begin{figure}[t]
  \centering
  \includegraphics[width=\linewidth]{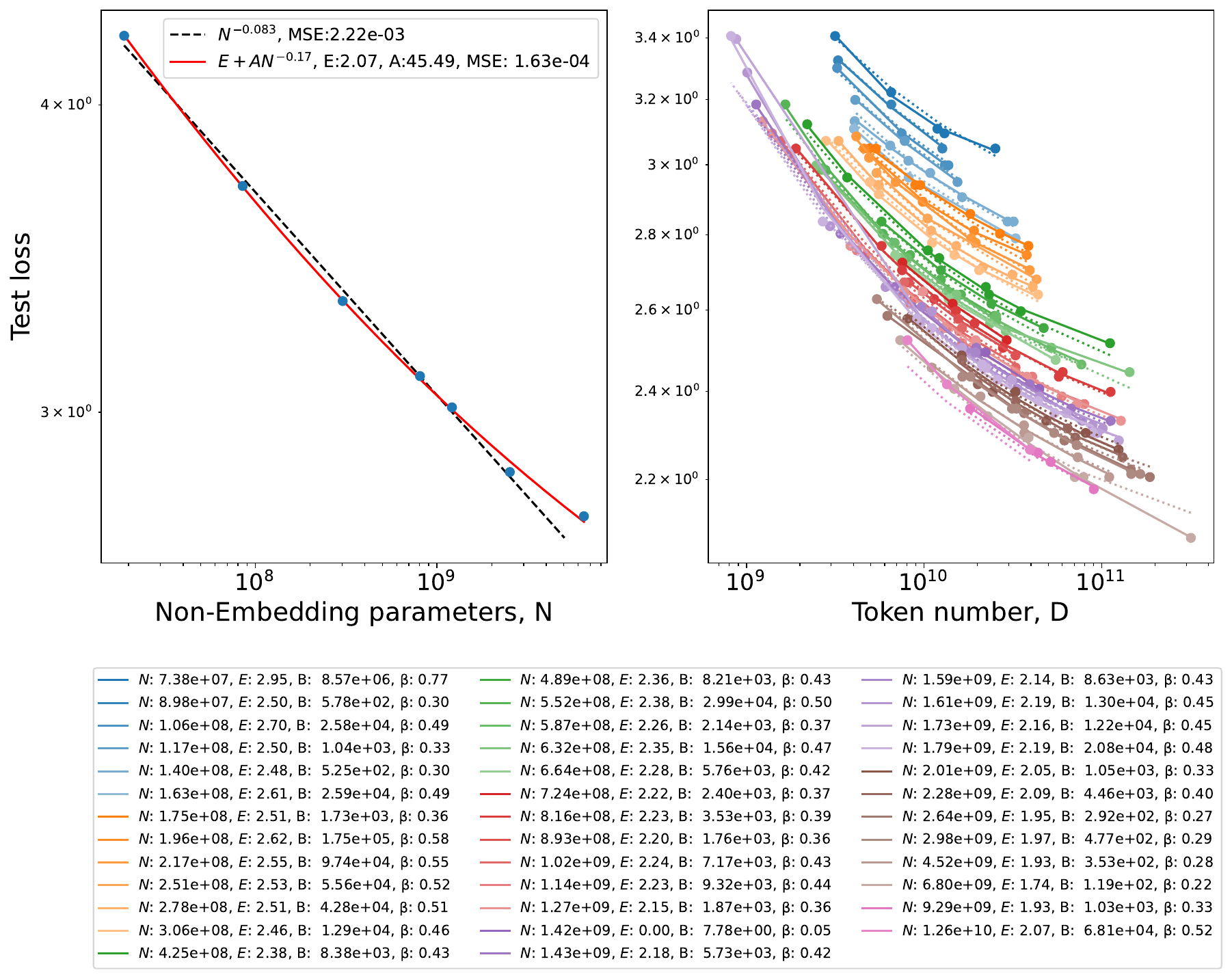}
  \caption{Left: Fits for $L(N)_D$ for the Pythia models. Right: 1d $L(D)_N$ fits for Chinchilla data (solid lines) with parameters shown in the legend, $L(N,D)$ fits to 2d Chinchilla formula (dotted lines). $N$ includes embedding parameters. \label{fig:pythia_chinchilla}}
\end{figure}

An important upshot of this discussion is that while it is common to present power-law scaling as a linear fit to the loss on a log-log plot, more accurately the fit should be curved. One could attempt to fit a line to the reducible loss, $L - E$, however this requires an extremely accurate estimate of $E$, particularly in the tails of the power law, which is often not available without first finding the best fit for $L$. 

\subsection{2d Chinchilla fit is relatively poor}
\label{Sec:chinchilla_analysis}

Next, to address point (2) above, we revisit the Chinchilla results \citep{hoffmann2022training}. The raw data for $L(N,D)$ reported in \citep{hoffmann2022training} was extracted and shared in \citep{besiroglu2024chinchillascalingreplicationattempt}, which we used for our analysis below.  First, we compare three different methods for fitting a two-dimensional function $L(N,D)$, which allows determination of the compute optimal scaling laws: (i) fitting to the 2d Chinchilla formula, Eq. \ref{2dChinchillaEq}, (ii) neural network regression, and (iii) kernel regression. We refer to Appendix \ref{app:fitting} for details regarding the neural network and kernel fits. The raw Chinchilla data contains $245$ pairs of $(N,D)$ values. Following \citep{besiroglu2024chinchillascalingreplicationattempt}, we throw out the datapoints corresponding to the 5 largest values of $L(N,D)$, which appear to be outliers due to poor optimization. 

\begin{figure}[H]
  \centering
  \includegraphics[width=4in]{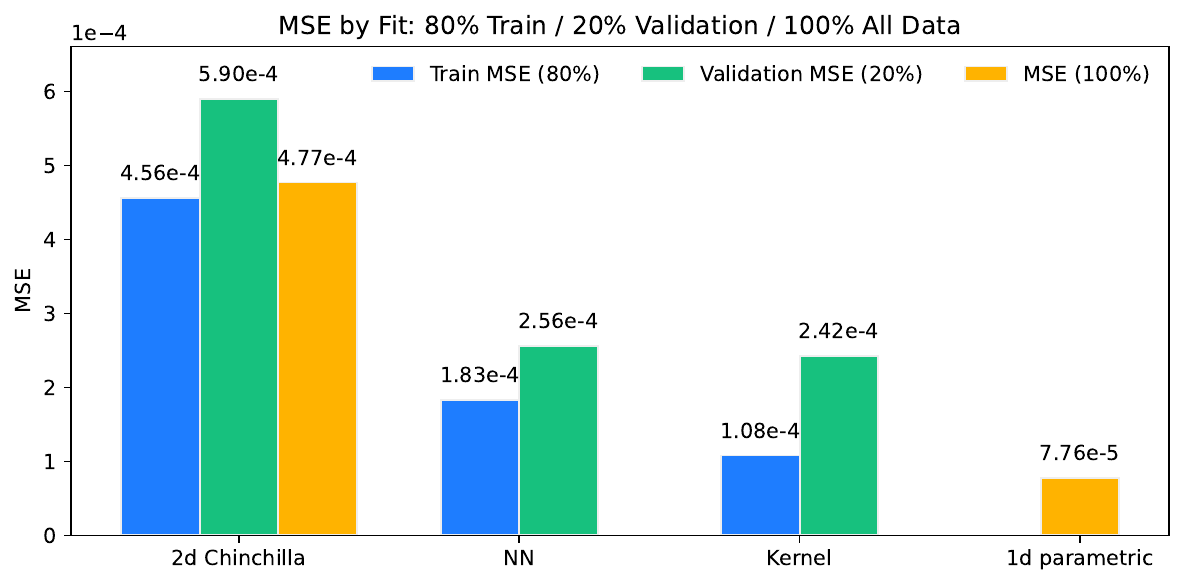}
  \caption{\label{fig:chinchilla_mse_bar} Train and validation MSEs for the 2d Chinchilla fit, neural network regression, kernel regression, and the 1d parametric fit of Eq. \ref{1dlossfit}, for the Chinchilla data \citep{hoffmann2022training,besiroglu2024chinchillascalingreplicationattempt}. }
\end{figure}

In all three cases, we perform an 80/20 train/validation split on the data, fit using a Huber loss ($\delta = 10^{-3}$) as in \citep{hoffmann2022training}, and report the MSE test loss averaged over 20 random splits. We note that such classical machine learning methods are not typically reported in the contemporary neural scaling law literature. The train / validation MSEs, averaged over the 20 train / validation splits, are displayed in Fig. \ref{fig:chinchilla_mse_bar}. We see that the neural network and kernel fits gives an almost $2\times$ factor improvement in the average validation MSE. From a model selection perspective, one should use the kernel or neural network fit over the 2d Chinchilla fit. 

The drawback of the neural network and kernel fits are that they are not interpretable. To also have interpretable conclusions, we consider the one-dimensional fit $L(D)_N$ given in Eq. \ref{1dlossfit}. The result is plotted in Fig. \ref{fig:pythia_chinchilla} and the MSE is shown in Fig. \ref{fig:chinchilla_mse_bar}. The 1d equations give a significantly better fit, with a MSE of $7.76\times 10^{-5}$. This is almost an order of magnitude better than the 2d fit, which gives an MSE of $4.77 \times 10^{-4}$. This is unsurprising given that there are many more parameters, but also indicates that it is more accurate to consider an exponent $\beta_N$ varying with $N$. 

The 2d fit yields the parameters  $A =477.82$, $B = 2143.62$, $E = 1.8172$, $\alpha = 0.3473$, $\beta = 0.3672$. The 1d fit, $L(D)_N = E_N + B D^{-\beta_N}$ gives a range of exponents, $\beta_N \sim 0.3 - 0.6$, albeit with some outliers; specifically, mean $\overline{\beta_N} = 0.41$ and standard deviation $0.085$ after throwing out the two extreme outliers. While these are consistent with each other, we see $\beta_N$ does have considerable variation with $N$. There are not enough datapoints for fixed $D$ to obtain a fit for $L(N)_D$. 

One can also use the neural network or kernel fits to extract compute optimal curves, using the methodology described in Sec. \ref{sec:methodology}. In Appendix \ref{app:compute_optimal} Fig. \ref{fig:chinchilla-nn-compute-optimal}, we present the results for compute optimal curves computed using the neural network fit to $L(N,D)$. Our method yields the exponents $\gamma = 0.16$, $a = 0.482$, $b = 0.504$. These results are in relatively good agreement with the 2d Chinchilla fit ($\gamma = \frac{\alpha \beta}{\alpha + \beta} = 0.178$, $a = \frac{\beta}{\alpha + \beta} = 0.513$, $b = \frac{\alpha}{\alpha + \beta} = 0.487$). 

\subsection{Recovering basic scaling law results with 2 layer transformers and short context lengths}

\begin{figure}[t]
  \centering
  \includegraphics[width=\linewidth]{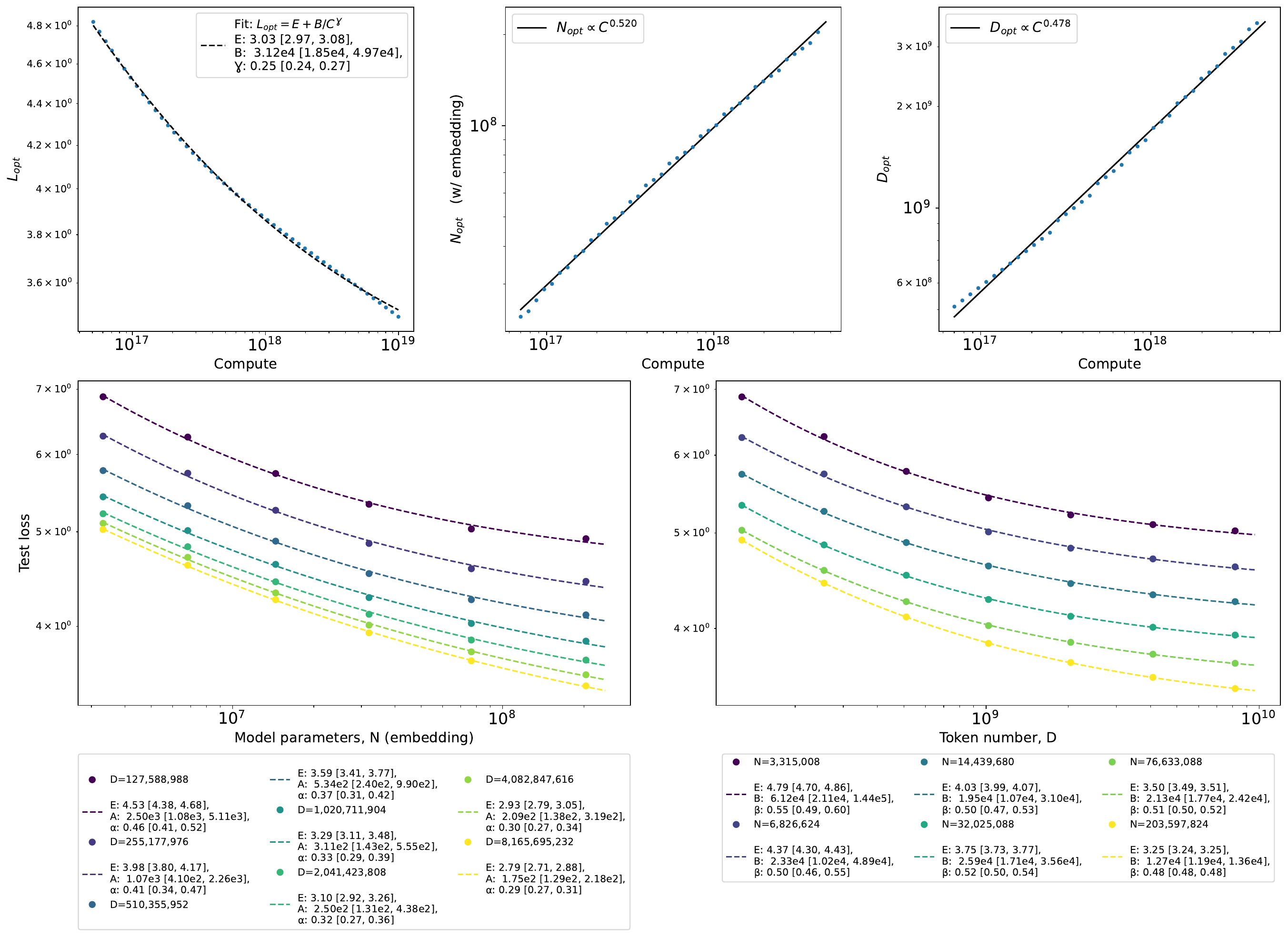}
  \caption{2 layer transformer trained on language with $S P$, where $N$ includes the embedding parameters. Mean exponent $\overline{\alpha_D} = 0.354$ with standard deviation $0.058$. Mean exponent $\overline{\beta_N} = 0.511$ with standard deviation $0.022$. Average MSE for $L(N)_D$ 1d power law fits is $5.41 \times 10^{-4}$. Average MSE for $L(N)_D$ 1d exp fits: $6.09 \times 10^{-3}$. Average MSE for $L(D)_N$ 1d power law fits: $1.88 \times 10^{-4}$. Average MSE for $L(D)_N$ 1d exponential fits: $1.00 \times 10^{-2}$. 
}
  \label{fig:language_sp_embed}
\end{figure}

Here we discuss the findings summarized in point (3) at the beginning of this section. We train a 2 layer transformer with context length $100$, using a fixed batch size of $512$ sequences, with standard parameterization. We use embedding dimensions $n_{\text{embd}} = 64, 128, 256, 512, 1024, 2048$. The number of attention heads is $\text{max}(4, n_{\text{embd}}/64)$. 

\begin{figure}[t]
  \centering
  \includegraphics[width=\linewidth]{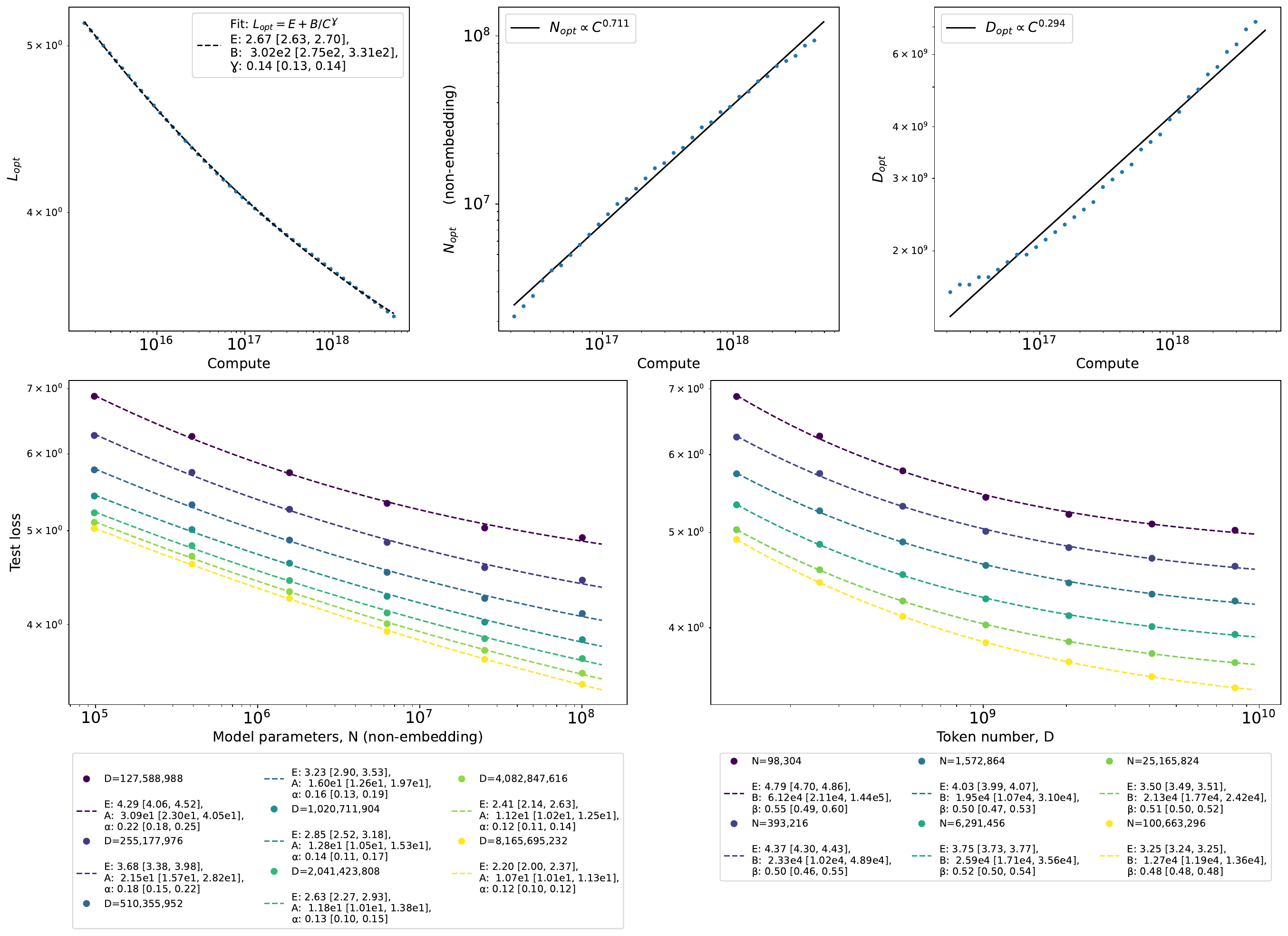}
  \caption{2 layer transformer trained on language with $S P$. Here $N$ refers to non-embedding parameters, in contrast to the rest of the paper. }
  \label{fig:language_sp_nonembed}  
\end{figure}

\begin{figure}[tb]
  \centering
  \includegraphics[width=\linewidth]{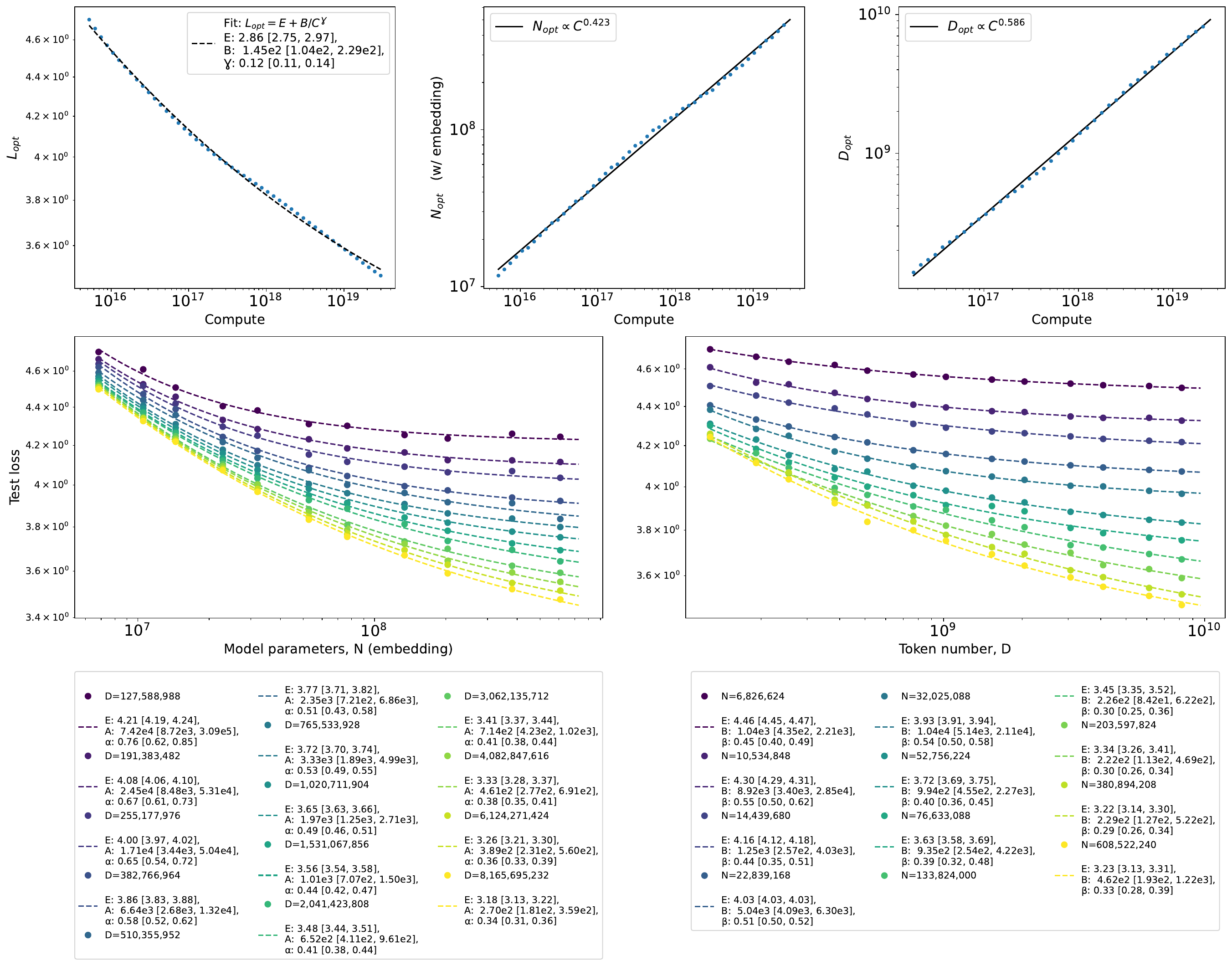}
  \caption{2 layer transformer trained on language (Fineweb-edu) with $\mu P$, context length $50$, and batch size of $100$ sequences. Mean exponent $\overline{\alpha_D} = 0.501$ with a standard deviation of $0.125$. Mean exponent $\overline{\beta_N} = 0.408$ with standard deviation $0.093$. Average MSE for $L(N)_D$ 1d power law fits is $1.19 \times 10^{-4}$, compared to $1.76 \times 10^{-3}$ for the best exponential fits. Average MSE for $L(D)_N$ 1d power law fits is $9.05 \times 10^{-5}$, compared to  $1.32 \times 10^{-3}$ for the best exponential fits. 
  \label{fig:2layer_muP_language}  }
\end{figure}

\begin{figure}[tb]
  \centering
  \includegraphics[width=\linewidth]{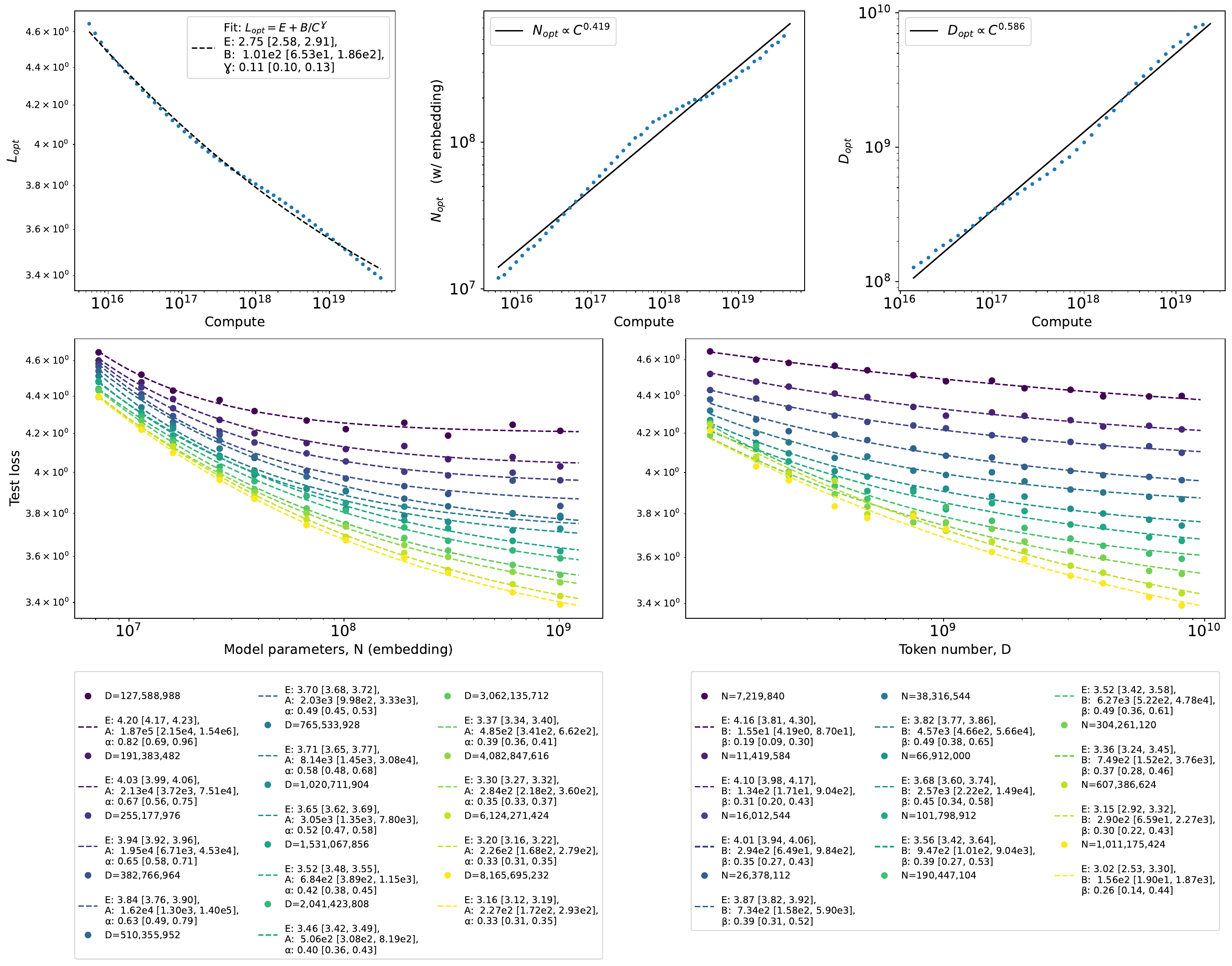}
  \caption{4 layer transformer trained on language (Fineweb-edu) with $\mu P$, context length $50$, and batch size of 100 sequences. Mean exponent $\overline{\alpha_D} = 0.506$ with a standard deviation of $0.148$. Mean exponent $\overline{\beta_N} = 0.362$ with standard deviation $0.090$. Average MSE for $L(N)_D$ 1d power law fits is $2.54 \times 10^{-4}$, compared to $2.30 \times 10^{-3}$ for the best exponential fits. Average MSE for $L(D)_N$ 1d power law fits is $3.58 \times 10^{-4}$, compared to  $2.04 \times 10^{-3}$ for the best exponential fits. }
  \label{fig:4layer_muP_language}  
\end{figure}

The 1d fits for $L(N)_D$ and $L(D)_N$, where $N$ either includes embedding parameters or not, are shown in Figs. \ref{fig:language_sp_embed}, \ref{fig:language_sp_nonembed}. The $\alpha_D$ are in the range $\alpha_D \sim 0.29 - 0.46$, with a mean of $\overline{\alpha_D} = 0.354$, which is close to the result for $\alpha$ from the 2d Chinchilla fit. The $\beta_N$ are in the range $0.48 - 0.55$, with a mean $\overline{\beta_N} = 0.511$. These are slightly higher, but within the same range, than the results in the preceding section where we refit the Chinchilla data.

The 2D Chinchilla fit yields the parameters 
$A \approx 507$, $B=20094$, $E=2.802$, $\alpha=0.365$, $\beta=0.503$. The exponents $\alpha$, $\beta$ are close to the means $\overline{\alpha_D}$, $\overline{\beta_N}$ obtained from the 1d fits. These imply the compute optimal exponents $L_{\text{opt}} \sim C^{\gamma}$ with $\gamma = \frac{\alpha \beta}{\alpha + \beta} = 0.212$. In Fig. \ref{fig:language_sp_embed}, we present the compute optimal scaling curves obtained using the neural network fit (including embedding parameters), which is within $20\%$ of the prediction from the 2d Chinchilla formula. 

Figs. \ref{fig:language_sp_embed}, \ref{fig:language_sp_nonembed}  also depict the scaling exponents for the optimal parameter and data size as a function of compute, $N_{\text{opt}} \sim C^a$, $D_{\text{opt}} \sim C^b$, for both the case where embedding parameters are and are not included. From the fits, we see that when embedding parameters are included, we have $a/b \approx 0.94$, which is close to the famous Chinchilla result that $a/b \approx 1$. When embedding parameters are not included, we have $(a,b) = (0.742, 0.386)$, which is close to the OpenAI result of $(0.73, 0.27)$ \citep{kaplan2020scaling}. 
Our result is in line with observations in \citep{porian2024resolving,pearce2024reconciling}, that a major part of the discrepancy between the Chinchilla and OpenAI scaling laws is whether embedding parameters are included. These prior works also identified learning rate schedules and other factors in contributing to the discrepancy; in light of our results, these might be more relevant for deeper models. 

\subsection{Scaling laws with $\mu$P: improved parameter efficiency?}

Here we present our results for 2 and 4 layer transformers trained on Fineweb-edu, with the same experimental settings as the main body of the paper: maximal update ($\mu P$) parameterization, context length of $50$, and a fixed batch size of $100$ sequences. The results are qualitatively similar to results presented in other cases. The main interesting point is that we find that for both $2$ and $4$ layer transformers, the mean exponent $\overline{\alpha_D} \approx 0.5$, which is larger than the case of SP, which had $\overline{\alpha_D} \approx 0.35$. This indicates that these settings are more parameter efficient, as the loss decreases more sharply with $N$. This is also reflected in the compute optimal exponents $(a,b) \approx (0.42, 0.58)$, which shows that the optimal parameter size need not grow as quickly with compute as in the case of standard parameterization; that is, compute optimal training is not at fixed token number per parameter. Since $\mu P$ was designed to give $O(1)$ feature updates in the large width limit \cite{yang2021tensor} it is plausible that it could be more parameter efficient for compute optimal training. It would be interesting to see if these results hold robustly at larger scales. 

\section{Additional related work}

\noindent {\bf Transformers and random walks / Markov chains.} \citep{Perozzi_2014,shi2025towards,makkuva2025attention,edelman2024evolution} study the ability of neural networks to learn random walks or  Markov processes, often focusing on model capability and learnability. Our work instead focuses on using random walks on graphs as a paradigm for understanding neural scaling laws. 

\noindent{\bf Neural scaling laws} Several works in addition to those referenced earlier also examine the origin of neural scaling laws. \citep{sharma2022scaling,bahri2021explaining} show that the existence of a data manifold implies neural scaling laws. The data in our work consists of sequences of discrete tokens, which have no obvious data manifold, so such an explanation appears to be inapplicable. 

Other mechanisms for neural scaling laws that have been studied include 
\citep{liu2025superposition}, which isolates ``superposition" as an important underlying principle, and  \citep{song2024resourcemodelneuralscaling} which proposes a resource-based theory. It would be interesting if these perspectives could be related to scaling laws we observed in the context of random walks. \citep{hutter2021learningcurvetheory} was an early paper suggesting that power laws in the data were responsible for ``interesting" scaling laws (i.e. data exponents $\alpha \neq 0.5$ or $1$) with respect to dataset size, although there is no neural network architecture and associated model parameter scaling law in Hutter's model.

\noindent{\bf Scaling laws with $\mu P$} \citep{dey2023cerebras} studied compute-optimal scaling in the case of fixed tokens-per-parameter, finding that $\mu P$ gave $\sim 1 \%$ better loss and more robust scaling laws. We are not aware of reports in the literature of the scaling law exponents for $\mu P$; in this paper we find that $\mu P$ is more parameter-efficient (larger $\overline{\alpha_D}$ and smaller $a$) than SP at relatively small scales. 

\section{Discussion}

We have demonstrated that neural scaling laws, as observed empirically in large language models, can also be observed in the simpler setting of Markov random walks on graphs. We have studied a wide variety of settings: (i) Erdös-Renyi graphs (1K nodes, 5K edges; 8K nodes, 50K edges; 50K nodes, 2M edges) with unbiased ($\kappa = 0$) and biased ($\kappa = 1$) random walks; (ii) scale free Barabási-Albert graphs (8K nodes, 50K edges); (iii) language bigram graphs (50K nodes, 33M edges). We have also demonstrated that one can consider datasets of monotonically increasing complexity, from language bigrams, TnL, to natural language data, and that one can track the mean scaling exponents as the complexity is increased, revealing a monotonic decrease in $\overline{\alpha_D}$. 

Upon analyzing the scaling exponents, we see that most of them appear to the right of the $\overline{\alpha_D} = \overline{\beta_N}$ line in Fig. \ref{fig:dialing_exponents}. With some small deviations, we generally have $\overline{\alpha_D} \gtrsim \overline{\beta_N}$, which suggests that the loss is limited more by data than by architectural capacity. This may be related to our setting of 1 epoch training, where the number of training steps is tied to the dataset size; it is possible that relaxing this would make the models more sample efficient and increase $\overline{\beta_N}$. It is also likely related to the fact that the setting of Markov random walks is too simple to be limited by architectural capacity. It would be interesting to find settings that are more limited by architectural capacity than by dataset size, so that $\overline{\beta_N} \gg \overline{\alpha_D}$; this would be expected in more complex datasets, where larger architectures are required to effectively learn, but where data is used more effectively, for example through multi-epoch training, data augmentation, synthetic data, self-play, and so on. 

One of the key points of our paper, as exhibited by our results on unbiased random walks on ER graphs (Fig. \ref{fig:ER8k}), is that empirically neural scaling laws arise even when the data itself has no power law correlations. This calls for alternative explanations and a deeper understanding of neural scaling laws than have been provided so far in the literature. In the setting of one epoch training for a fixed dataset size $D$, the number of optimization steps is also fixed. As we increase $N$ with fixed $D$, the scaling laws tell us that the model is able to reach a lower loss with the same number of training steps. Equivalently, the model is able to reach a given value of the loss with fewer steps. This suggests that the basin of low loss solutions is opening up in a particular way with model size, which allows gradient-based optimization to be more efficient. A potential path forward could be to study how the loss landscape is opening up with $N$ and how this affects the gradients as a function of $N$ during training. 

In the kernel regression models referenced earlier, power law scaling in the loss arose from power laws in the features. While we have examples where the data has no power laws, it might be the case that power laws develop in the internal activations of the model, which then drive power law scaling in the loss. It would be interesting to further study these internal activations and see to what extent signatures of power laws arise. 

While we found one method to dial the complexity of the data from bigrams to natural language via TnL, it would be interesting to develop alternative more controllable and systematic ways of interpolating in complexity between bigrams and natural language, such as through $n$-grams, skip $n$-grams, and hierarchical graphical structures. We leave a detailed study of these directions for future work. 

Finally, we note that a difficult and important issue with scaling laws is to understand their robustness. First, the optimization algorithm is an important part of the story. All scaling laws observed empirically depend on a grid search over hyperparameters for a particular choice of optimization algorithm. Improving the optimization algorithm and the hyperparameter search will lower the loss $L(N,D)$; if different model sizes respond differently to these choices, the scaling laws could be altered substantially. In principle, a minimization over all possible optimization algorithms and hyperparameters separately at each $N$ might remove the power law scaling in $L(N)_D$ altogether, because in many cases the shortcoming of smaller models is not expressivity but rather learnability. 

Second, we observed, even in the case of Chinchilla's data, that the parameters of the best fits were rather sensitive to the window over which the fitting was performed. As detailed in Appendix \ref{app:fitting}, this is different from the bootstrap confidence intervals reported in our fits. A detailed sensitivity analysis is important for practitioners, and we leave it to for a separate careful study. Here we note that in addition to the exponents themselves, the sensitivity of the exponents is also an important quantity to characterize. Complex datasets like natural language might have more robust exponents than random walks; the former requires learning substantial interrelated hierarchical structures, while the former is more about learning a (large) table of transition probabilities. In our analysis, we saw some preliminary indications of this -- our random graph results depended on careful hyperparameter grid searches, whereas our natural language results appeared to be more robust. We expect that further research along these lines will reveal additional insights into the origin of neural scaling laws. 

\section{Acknowledgments}

We thank Newsha Ardalani, Yasaman Bahri, and Andrew Cohen for discussions and Rylan Schaeffer, Dan Roberts, and Cengiz Pehlevan for comments on the draft. MB thanks the Simons Collaboration on Physics of Learning and Neural Computation (SFI-MPS-POL-00012574-09) for support while at UMD. 

\appendix

\section{Experimental details}
\label{app:exp_details}

\noindent\textbf{Transformer architectures.}
We train pre-normalization decoder-only transformer models \citep{vaswani2017attention,radford2019language} using rotary positional embedding \citep{black2022gpt}. We follow the rotary implementation available on Huggingface, with the default frequency base of 10,000 and with rotary dimension equal to the embedding dimension. The model vocabulary size is $V$, where $V$ is the number of nodes in the graph. The models have 2 or 4 hidden layers, with embedding dimension ranging from $n_{\text{embd}} = 128$ to $4096$. For each model, we set the number of attention heads to $\text{max}(4, n_\text{embd}/64)$. No tokenizer is required, as each node in our graphs is treated as a distinct token with its own unique token ID. We tie the token embedding matrix and the output head weights \citep{press2017using}.

\noindent\textbf{Optimization.}
We initialize all models from scratch and pre-train them by minimizing the cross-entropy loss between the ground-truth data and predicted sequences. Model parameters are optimized using AdamW with a weight decay of 0.01 and a peak learning rate $\lambda$. We sweep $\lambda$ over a grid of 14 values ranging from $5 \cdot 10^{-6}$ to $0.1$. We use a learning rate schedule with a linear warmup for the first 2\% of training steps, followed by cosine decay to 0. On each training step, we stack 100 independent random-walks into a tensor of shape (100, $T$). We use the parameter initialization scheme and $\mu$P implementation from the nanoGPT-muP repository \citep{mup_repo}; we found that $\mu P$ helped mitigate training instabilities across the sweep of leaning rates for our random walk experiments. For each hyperparameter configuration, we train 3 models with different random initialization seeds and we report the best test loss performance. 

\noindent\textbf{Data generation.}
We introduce synthetic random-walk datasets. We generate sequences using the following procedure:

\begin{itemize}
    \item For a desired sequence length $T$ (generally 50 or 100), we start from a weighted transition matrix $W_{uv}$ and an initial-node distribution $M_u$ where $\sum_{v} W_{uv} = 1$ for every node $u$ and $\sum_{u} M_u = 1$. $W_{uv}$ is the probability of moving from node $u$ to node $v$ in one step and $M_u$ is the probability of choosing node $u$ as the starting point.
    \item We sample the first node of each sequence from $M_u$.
    \item We then sample the remaining $T-1$ nodes by repeatedly drawing the next node $v$ according to $W_{uv}$, where $u$ is the current node.
    \item These walks may revisit nodes and traverse the same edge multiple times (i.e. node and edge repetition is allowed).
\end{itemize}

For Erdös-Renyi or Barabási-Albert graphs, we first create a graph $\mathcal G$ with $V$ nodes and $E$ edges, yielding a binary symmetric adjacency matrix $A_{uv}$. For biased walks, we sample $W^{init}_{uv}$ from a power-law distribution $\text{Pr}(W^{init}_{uv} = k) \propto k^{-\kappa}$, for $k \in [k_{\text{min}}, k_{\text{max}}]$ whenever $A_{uv} = 1$ and set $W^{init}_{uv} = 0$ otherwise. For unbiased walks, we let $W^{init}_{uv} = A_{uv}$. 

For bigram graphs, we first tokenize the Fineweb-edu-10B dataset \citep{penedo2024fineweb} using the GPT-2 tokenizer \citep{radford2019language}, which has a vocabulary of size $50,257$. This tokenized dataset has approximately $130$M unique bigrams. We remove all bigrams that appear $5$ times or fewer, leaving around $33$M unique bigrams. Let $C_{uv}$ denote the number of occurrences of the bigram $(u, v)$ in the filtered corpus and set $W^{init}_{uv} = C_{uv}$.

Finally, we define $$W_{uv} = \frac{W^{init}_{uv}}{\sum_{j} W^{init}_{uj}} \quad \mathrm{and} \quad M_u = \frac{\sum_{j} W^{init}_{uj}}{\sum_{i,j} W^{init}_{ij}}.$$

For T1L, T2L and T4L, we train a $n$-layer transformer (respectively $n=1,2,4$) with an embedding dimension of 4096 and a vocabulary size of 50,257 on the Fineweb-edu-10B dataset \citep{penedo2024fineweb}. We train with a context length of 50, batch size of 100 and choose the learning rate $\lambda$ that minimizes the training loss. After training, we sample random sequences from the model using temperature $1$.

All natural language experiments are done with Fineweb-edu.

\section{Fitting scaling laws}
\label{app:fitting}

\subsection{$L(N)_D$ and $L(D)_N$ fits}

To fit $L(D)_N$ and $L(N)_D$, we fit the power law
\[
y(x)= E + B\,x^{-\beta}\quad (\beta>0)
\]
by minimizing a Huber-robust least-squares objective with cutoff $\delta$. Specifically, for residuals
$r_i=y(x_i;\theta)-y_i$, we minimize $\sum_i \rho_\delta(r_i)$, where
\[
\rho_\delta(r)=
\begin{cases}
\frac12 r^2,& |r|\le \delta,\\[4pt]
\delta\left(|r|-\frac12\delta\right),& |r|>\delta,
\end{cases}
\]
so errors smaller than $\delta$ are treated quadratically while larger errors are penalized approximately
linearly (down-weighting outliers). We set $\delta = 1.4826 \text{MAD}(y)$, where $\text{MAD}(y) = \,\mathrm{median}\!\left(\left|y-\mathrm{median}(y)\right|\right)$ is the median absolute deviation, and provides an estimate for an appropriate scale of $\delta$. (If $y$ is pure Gaussian noise, this reduces to $\delta = \sigma$, the standard deviation of the Gaussian noise; on the other hand, if $y$ is dominated by signal and noise is small, $\delta$ is large and the fit reduces to ordinary least squares). We use a fallback $\delta=0.1\,\mathrm{std}(y)$ if the median absolute deviation is zero. For numerical stability we reparameterize as
\[
y(x)=E + A\left(\frac{x}{x_0}\right)^{-\beta},\qquad x_0=\mathrm{median}(x),\ A>0,
\]
optimize over $(E,\log A,\log\beta)$ with bounds $\beta\in[\beta_{\min},\beta_{\max}]$, and then report
$B=A\,x_0^{\beta}$. The bounded nonlinear optimization (trust-region reflective algorithm) is run from 40 initializations: $E_0$ is chosen from $y$-quantiles, with $(A_0,\beta_0)$ obtained from an approximate log-linear fit of $\log(y-E_0)$ versus $\log(x/x_0)$ on points where $y-E_0>0$, plus additional random starts; we keep the solution with the smallest final Huber objective.

\begin{figure}[t]
  \centering
  \includegraphics[width=0.5\linewidth]{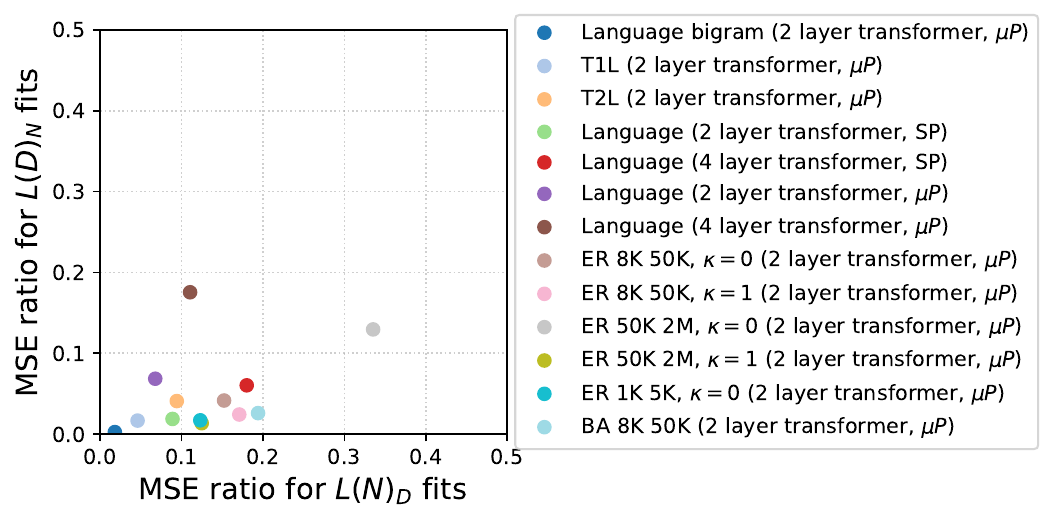}
  \caption{\label{MSE_ratios} Ratio of the mean-squared error between the best power-law and the best exponential ($y = a + be^{-cx}$ fits, in order to validate the power-law scaling hypothesis. The power law fits for $L(D)_N$ are extremely good. The ones for $L(N)_D$ are generally not as clean, but it is unmabiguous that the power law form gives a substantially better fit. Biased random walks with power-law weights give clearer power-laws than unbiased random walks for Erdös-Renyi graphs.}
  \label{fig:mse_ratios}
\end{figure}

For all of our $L(N)_D$ and $L(D)_N$ fits, we compare the quality of the power law fit to the quality of the best exponential fit, $y = a + b e^{-cx}$, by examining the MSEs of the power law and exponential fits. The ratios of the power law and exponential MSEs for the datasets studied in this paper are shown in Fig. \ref{fig:mse_ratios}. This demonstrates that the power law fits for $L(D)_N$ are extremely good (a factor of 50-100x better than the best exponential fits). The power law fits for $L(N)_D$ are also good; with one exception, they are between 5-100x better than the best exponential fits. 

\subsection{Confidence intervals}

In our fits we report 95\% confidence intervals (shown in square brackets in the figure legends) for $\theta=(E,B,\beta)$ in the power law $y(x)=E+B\,x^{-\beta}$ using a fixed-$x$ wild bootstrap with BCa correction. After obtaining the robust point estimate $\hat\theta$ and fitted values $\hat y_i=\hat E+\hat B x_i^{-\hat\beta}$, we form residuals $r_i=y_i-\hat y_i$. For each bootstrap replicate $b=1,\dots,N_{\mathrm{boot}}$, we generate
\begin{equation}
y_i^{*(b)}=\hat y_i+\varepsilon_i^{(b)}\,r_i,
\end{equation}
with i.i.d.\ weights $\varepsilon_i^{(b)} = \pm 1$ satisfying $\mathbb{E}[\varepsilon]=0$ and $\mathrm{Var}(\varepsilon)=1$. We refit the model to $\{(x_i,y_i^{*(b)})\}$ using the same Huber objective and the same cutoff $\delta$ as in the base fit, warm-starting the optimizer from the base solution; failed refits are discarded.

For each scalar parameter $t\in\{E,B,\beta\}$ we obtain bootstrap draws $\{\hat t^{*(b)}\}_{b=1}^{N_{\mathrm{boot}}}$ and use BCa (bias-corrected and accelerated) intervals. The \emph{bias-correction} term is
\begin{equation}
z_0 \;=\; \Phi^{-1}\!\left(\frac{\#\{\hat t^{*(b)}<\hat t\}+ \tfrac12\,\#\{\hat t^{*(b)}=\hat t\}}{N_{\mathrm{boot}}}\right),
\end{equation}
which measures how the bootstrap distribution is shifted relative to the point estimate $\hat t$ (here $\Phi$ is the standard normal CDF). The \emph{acceleration} term $a$ is estimated via a leave-one-out jackknife: for each $i=1,\dots,n$ we refit the model on the dataset with observation $i$ removed, yielding jackknife estimates $t_{(i)}$ and mean $\bar t_{(\cdot)}=\frac1n\sum_i t_{(i)}$. We then set
\begin{equation}
a \;=\; \frac{\sum_{i=1}^n\left(\bar t_{(\cdot)}-t_{(i)}\right)^3}
{6\left[\sum_{i=1}^n\left(\bar t_{(\cdot)}-t_{(i)}\right)^2\right]^{3/2}}\, .
\end{equation}
Given nominal quantiles $\alpha/2$ and $1-\alpha/2$ (with $\alpha=0.05$ for 95\% CIs), we compute adjusted quantile levels
\begin{equation}
\alpha_{\mathrm{BCa}}(q)
\;=\;
\Phi\!\left(
z_0 + \frac{z_0+\Phi^{-1}(q)}{1-a\,(z_0+\Phi^{-1}(q))}
\right),
\end{equation}
and take the endpoints as the empirical $\alpha_{\mathrm{BCa}}(\alpha/2)$ and $\alpha_{\mathrm{BCa}}(1-\alpha/2)$ quantiles of $\{\hat t^{*(b)}\}$. We also report bootstrap standard errors as the sample standard deviation of $\{\hat t^{*(b)}\}$.

For the $L(D)_N$ and $L(N)_D$, we use $N_{\text{boot}} = 4000$ bootstrap samples. 

\subsection{Fitting $L(N,D)$}

We compare three different methods for fitting a two-dimensional function $L(N,D)$, which allows determination of compute optimal scaling laws. In all cases below, we use the Huber loss, with $\delta = 1 \times 10^{-3}$ to match the setting in \citep{hoffmann2022training, besiroglu2024chinchillascalingreplicationattempt}.

\noindent\textbf{Fitting with 2d Chinchilla formula.}\; We fit to Eq. \ref{2dChinchillaEq} following the methodology of \citep{hoffmann2022training}. We use the code and data from the github repository of \citep{besiroglu2024chinchillascalingreplicationattempt}.

\noindent\textbf{Neural network regression with 3-layer FCN.} We fit $L(N,D)$ with a 3-layer fully connected neural network containing linear layers of dimension $2 \times n$, $n \times n$, and $n \times 1$ and GeLU activations. For fitting the Chinchilla data, we used a width $n = 512$; for the rest of the fits we use $n = 256$. The model achieving the best loss is used, within 5000 epochs of AdamW (weight decay $1e-4$) or until the loss has stopped improving by more than $1e-6$ for more than 200 steps. 

\noindent\textbf{Kernel ridge regression.} We fit $L(N,D)$ with the ANOVA-style RBF kernel
\[
k(x,x') \;=\; w_n \exp\!\Big(-\tfrac{(n-n')^2}{2\ell_n^2}\Big)
\;+\; w_d \exp\!\Big(-\tfrac{(d-d')^2}{2\ell_d^2}\Big)
\;+\; w_{nd}\exp\!\Big(-\tfrac{(n-n')^2+(d-d')^2}{2\ell_{nd}^2}\Big),
\]
with nonnegative amplitudes $w_{\{\cdot\}}$ and length-scales $\ell_{\{\cdot\}}$, where $x = (n,d) = (\log_{10} N, \log_{10} D)$. This kernel induces the function decomposition
$f(x)=f_n(n)+f_d(d)+f_{nd}(n,d)$, recovering a standard RBF when $w_n=w_d=0$. We set $w_{\{\cdot\}} = \ell_{\{\cdot\}} = 1$, although in principle a hyperparameter search can be carried out. The inputs $x$ are centered and normalized and the outputs are centered. The regression minimizes the objective $\sum_i \rho_\delta( y_i - f_i ) + (\lambda/2) \alpha^T K \alpha$,  with $f = K \alpha$, where $\rho_\delta$ is the Huber loss. This is solved by iteratively reweighted least squares (IRLS) with per-sample weights $w_i = 1$ if $|r_i| \leq \delta$ and $w_i = \delta/|r_i|$ otherwise, and a linear system at each step: $(W K + \lambda I) \alpha = W y_c$

\subsection{Fitting compute optimal scaling laws, $L_{\text{opt}}$, $N_{\text{opt}}$, $D_{\text{opt}}$}
\label{app:compute_optimal}

We use the fits $\hat{L}(N,D)$ obtained using the neural network regression approach. We pick a grid of 50 or 100 points between the minimum and maximum values of $N$ in our raw data, and similarly for $D$. We then determine the compute optimal loss in two ways, by $\text{min}_{N} \hat{L}(N, C/6N)$ and $\text{min}_D \hat{L}(C/6D, D)$. For a sufficiently dense grid and smooth enough fit $\hat{L}$, these two should coincide. This gives us an estimate of the compute optimal loss $L_{\text{opt}}(C)$. We fit $L_{\text{opt}}(C)$ to the power law form using the same methodology we used for $L(D)_N$ and $L(N)_D$, although we instead use 200 bootstrap samples for the confidence intervals since we observed the confidence intervals are already stable for those values of $N_{\text{boot}}$. 

Next, from the fit $\hat{L}(N,D)$ over the grid described above, we obtain $N_{\text{opt}}(C)$ and $D_{\text{opt}}(C)$ using Eq. \ref{compute_optimal_2}. We fit then do a standard linear regression to fit $\log N_{\text{opt}}(C)$ and $\log D_{\text{opt}}(C)$ to a line to extract the compute optimal exponents $a$ and $b$.  

In almost cases, we have to clip by hand the lower and upper values of $C$. One reason is that near the boundaries $C_{\text{min}} = 6 N_{\text{min}} D_{\text{min}}$ and $C_{\text{max}} = 6 N_{\text{max}} D_{\text{max}}$, the fit is less accurate and so $L_{\text{opt}}(C)$, $N_{\text{opt}}$, and $D_{\text{opt}}$ exhibit spurious behavior in these limits. Another reason is that in some cases, $N_{\text{opt}}$ and $D_{\text{opt}}$ are beyond the grid for $N$, $D$ we have sampled. 

We observed in practice that the neural network regression usually appears to give more sensible results for the compute optimal curves as compared with the kernel regression approach, so in all of our figures we report results using the neural network regression. We eschew the 2d Chinchilla fit for the neural network regression approach because it gives significantly better MSEs for the fits, as shown in Figs. \ref{fig:chinchilla_mse_bar}, \ref{fig:2layer_language_muP_mse_bar} ,\ref{fig:4layer_language_muP_mse_bar}, and explained in the main text. 

In nearly all cases, we found that the parameters of the fits for the scaling laws are sensitive to the window chosen for the fits. In principle the confidence intervals should also include a sensitivity analysis with respect to the window of the fit; we leave a detailed such analysis, which is of particularly important in for natural language data, for future work. 

\begin{figure}[t]
  \centering
  \includegraphics[width=\linewidth]{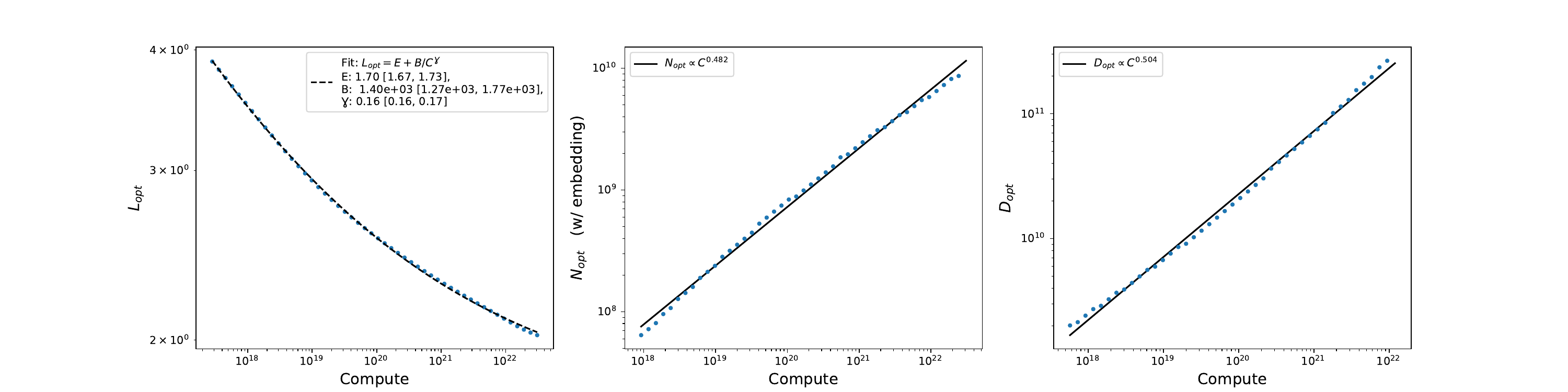}
  \caption{Compute optimal scaling laws obtained using the neural network fit for $L(N,D)$.}
  \label{fig:chinchilla-nn-compute-optimal}
\end{figure}

\begin{figure}[H]
  \centering
  \includegraphics[width=0.8\linewidth]{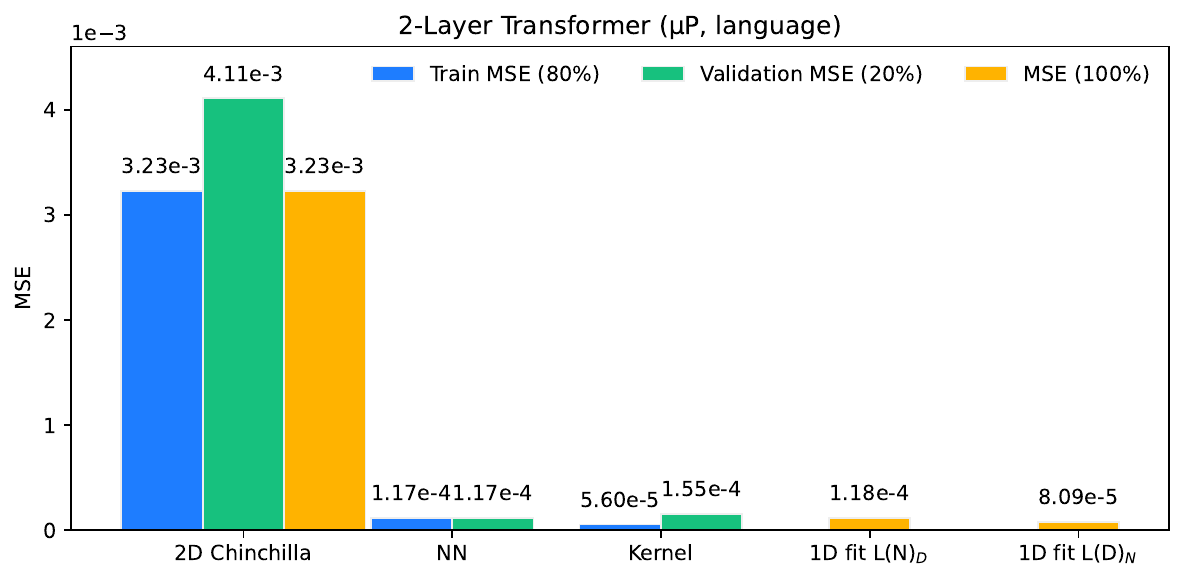}
  \caption{The 2d Chinchilla fit is around $30-40\times$ worse than the NN and kernel fits. $N$ = embedding parameters. Non-embedding is similar. 
}
  \label{fig:2layer_language_muP_mse_bar}
\end{figure}

\begin{figure}[H]
  \centering
  \includegraphics[width=0.8\linewidth]{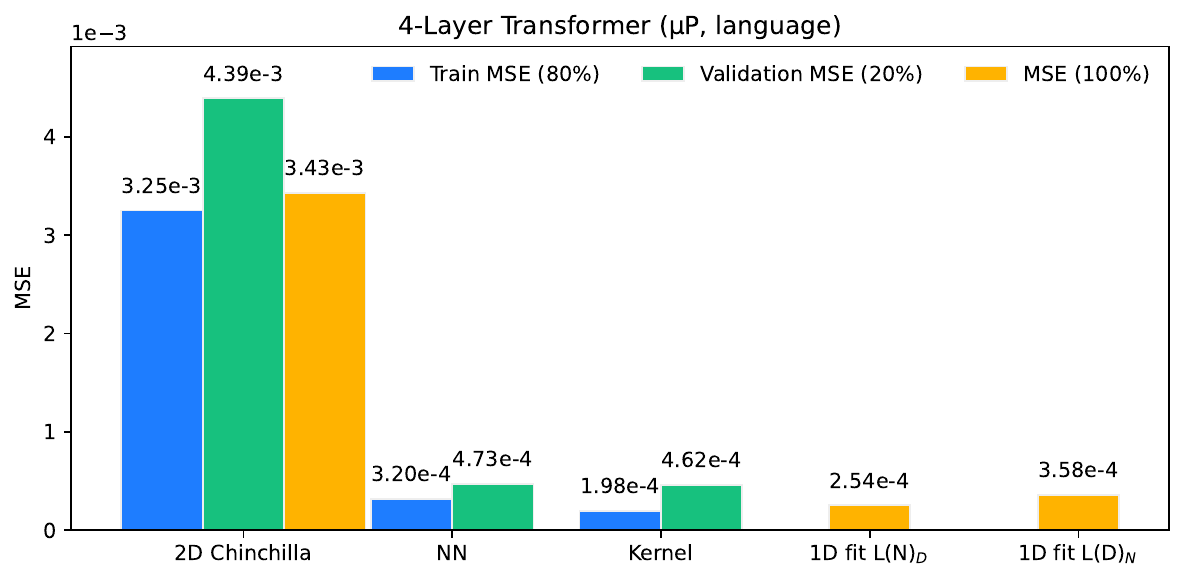}
  \caption{$N$ = embedding parameters. Non-embedding is similar. 2d Chinchilla fit yields the parameters $A=1838$, $B=677$, $E=3.342$, $\alpha=0.484$, $\beta=0.378$
}
  \label{fig:4layer_language_muP_mse_bar} 
\end{figure}

Fig. \ref{fig:chinchilla-nn-compute-optimal} demonstrates the neural network regression approach introduced above for obtaining compute optimal scaling curves for the raw Chinchilla data that we reanalyzed in Sec. \ref{Sec:chinchilla_analysis}. As described in the main text, the exponents are close to those obtained via other methods in \citep{hoffmann2022training,besiroglu2024chinchillascalingreplicationattempt}.

\section{Sample error baseline}
\label{app:baseline}

Consider the problem of estimating a probability distribution $\pi$ over a finite sample space $V$. Given a dataset of $D$ samples, consider the counting estimate 
\begin{align}
    \hat{\pi}_a = \frac{n_a}{D}, 
\end{align}
where $n_a$ is the number of times $a \in V$ appears in the dataset. Here we derive the data scaling laws for this counting model for cross-entropy and MSE losses, which provides a baseline against which to compare the random walk scaling law results. Note that this model was also studied in the Appendix of \citep{hestness2017deep} using a loss based on L1 norm, where a $1/\sqrt{D}$ scaling was computed; here we extend the analysis to MSE and cross-entropy loss, which gives a $1/D$ scaling law. 

If we let $X$ be a random variable with $V$ outcomes and probability distribution $\pi$, then $\hat{\pi}_a$ is the sample mean of the indicator function $I_a =\text{Ind}(X = a)$. The mean and variance of $\hat{\pi}_a$ are given by:
\begin{align}
\mathbb{E}[ \hat{\pi}_a ] &= \pi_a 
\nonumber \\
\text{Var}[ \hat{\pi}_a ] &= \mathbb{E}[(\hat{\pi}_a - \pi_a)^2] = \frac{1}{D} \pi_a ( 1- \pi_a)
\nonumber \\
\mathbb{E}[(\hat{\pi}_a - \pi_a)^3] &= 
\mathbb{E}[ \hat{\pi}_a^3 - 3 \hat{\pi}_a^2 \pi_a + 3\hat{\pi}_a \pi_a^2 - \pi_a^3] = \frac{1}{D^2} \pi_a (2 \pi_a -1)(\pi_a - 1)
\end{align}
Let us consider both mean-squared error and cross-entropy loss in this situation. By the central limit theorem, since $\hat{\pi}_a$ is the average over $D$ i.i.d. binary random variables, its distribution in the large $D$ limit is given by a Gaussian:
\begin{align}
    \hat{\pi}_a \sim \mathcal{N}(\pi_a, \frac{\sigma^2_a}{D}),
\end{align}
where $\sigma^2 = \pi_a (1 - \pi_a)$ is the variance of $I_a$. 

\subsection{Mean squared error}

\begin{align}
L_{MSE} = \frac{1}{|V|} \sum_{a\in V} \left(\pi_a - \hat{\pi}_a \right)^2 .
\end{align}
Its expected value is
\begin{align}
    \mathbb{E}[L_{MSE}] &= \frac{1}{|V|} \sum_{a\in V} \mathbb{E}[(\hat{\pi}_a - \mathbb{E}[\hat{\pi}_a])^2] 
    \nonumber \\
    &= \frac{1}{|V|} \sum_{a\in V} \text{Var}[\hat{\pi}_a]
    \nonumber \\
    &= \frac{1}{D} \frac{1}{|V|} \sum_{a\in V} \pi_a (1 - \pi_a)
\end{align}

\subsection{Cross-entropy loss}
The cross-entropy loss is \footnote{Note that for this to be well-defined for finite samples, where $\hat{\pi_a}$ might be zero even when $\pi_a \neq 0$, we must assume some smoothing, which we keep implicit.}
\begin{align}
    L_{CSE} = -\sum_{a \in V} \pi_a \log \hat{\pi}_a 
\end{align}
The expected value is:
\begin{align}
    \mathbb{E}[L_{CSE}] &= -\sum_{a \in V} \pi_a \mathbb{E}[ \log \pi_a (1 + \frac{(\hat{\pi}_a - \pi_a)}{\pi_a}) ]
    \nonumber \\
    &= S_\pi - \sum_{a \in V} \pi_a \sum_{n=1}^\infty \frac{(-1)^{(n+1)}}{n} \pi_a^{-n} \mathbb{E}[ (\hat{\pi}_a - \pi_a)^n],
\end{align}
where $S_\pi = - \sum_{a \in V} \pi_a \log \pi_a$ is the entropy of $\pi$. 
Evaluating the expectation values for $\hat{\pi}_a$ gives:
\begin{align}
    \mathbb{E}[L_{CSE}] &= S_\pi + \frac{1}{D} \frac{1}{2}\sum_{a \in V} (1 - \pi_a) - \frac{1}{3}\sum_{a} \frac{1}{\pi_a^2} \mathbb{E}[ (\hat{\pi}_a - \pi_a)^3] + \cdots
    \nonumber \\
    &= S_\pi + \frac{|V| - 1}{2D} - \frac{1}{3D^2} \sum_a \frac{(2\pi_a -1)(\pi_a - 1)}{\pi_a} + O(1/D^3)
    \nonumber \\
    &= S_\pi + \frac{|V| - 1}{2D} + \frac{1}{D^2} (|V| - \frac{2}{3} - \frac{1}{3}\sum_a 1/\pi_a)
     + O(1/D^3)    
\end{align}
We see that in both cases, we get an exponent $\beta = 1$ for the asymptotic power-law scaling of the population loss. These results can easily be verified empirically. 

\subsection{Application to graph random walk}

Let us consider applying the above to the graph random walk problem. There, for each node $v$, we want to learn a probability distribution $p(u | v)$ from the sequences. The node $v$ itself appears with probability $p(v)$. Therefore if we have a dataset of $D$ tokens, approximately $p(v) D$ of them will be $v$. So our effective dataset size for each node is $D_{\text{eff}, v}$, which is equal to $p(v) D$ in the large $D$ limit. For each node $v$ we have a counting estimate $\hat{p}(u|v)$, leading to a cross-entropy loss for each node:
\begin{align}
    \mathbb{E}[L_{CSE,v}] = S_{p(\cdot | v)} + \frac{\text{deg}(v) - 1}{2D_{\text{eff},v}} + O(1/D^2) .
\end{align}
The total cross-entropy loss, averaging over all nodes is then:
\begin{align}
    \mathbb{E}[L_{CSE}] &= \sum_v p(v) \mathbb{E}[L_{CSE,v}] = \sum_v p(v) S_{p(\cdot | v)} + \sum_v \frac{\text{deg}(v) - 1}{2D} + O(1/D^2)
    \nonumber \\
    &= \langle S_{p(\cdot | v)}\rangle + \frac{2E - n}{2D} + O(1/D^2)
\end{align}
For the unbiased random walk, $p(u|v) = A_{uv}/\text{deg}(v)$ and the stationary distribution on nodes is $p(v) = \text{deg}(v)/2E$. 

\section{Additional scaling law results}

\subsection{Number of bigrams vs. count: most bigrams appear only a few times}

\begin{figure}[H]
  \centering
  \includegraphics[width=0.5\linewidth]{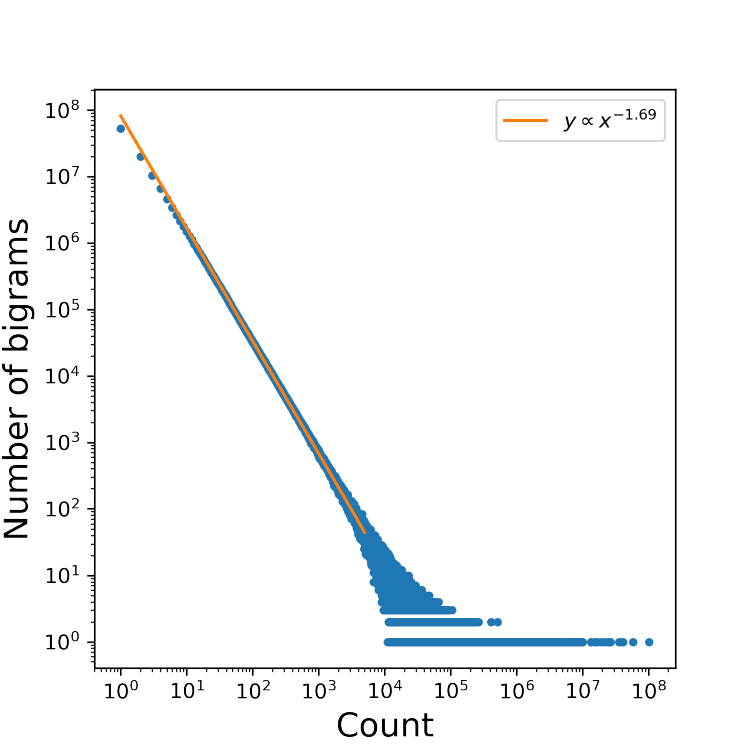}
  \caption{Number of bigrams that appear a given number (Count) of times, showing perfect linear fit on log-log scale. This plots demonstrates how almost all (over 100 million) bigrams appear only a handful of times. 
  \label{fig:bigram_stats}}
\end{figure}

\subsection{$4$ layer transformer trained on language (Fineweb-edu) with Standard Parameterization}

\begin{figure}[H]
  \centering
  \includegraphics[width=\linewidth]{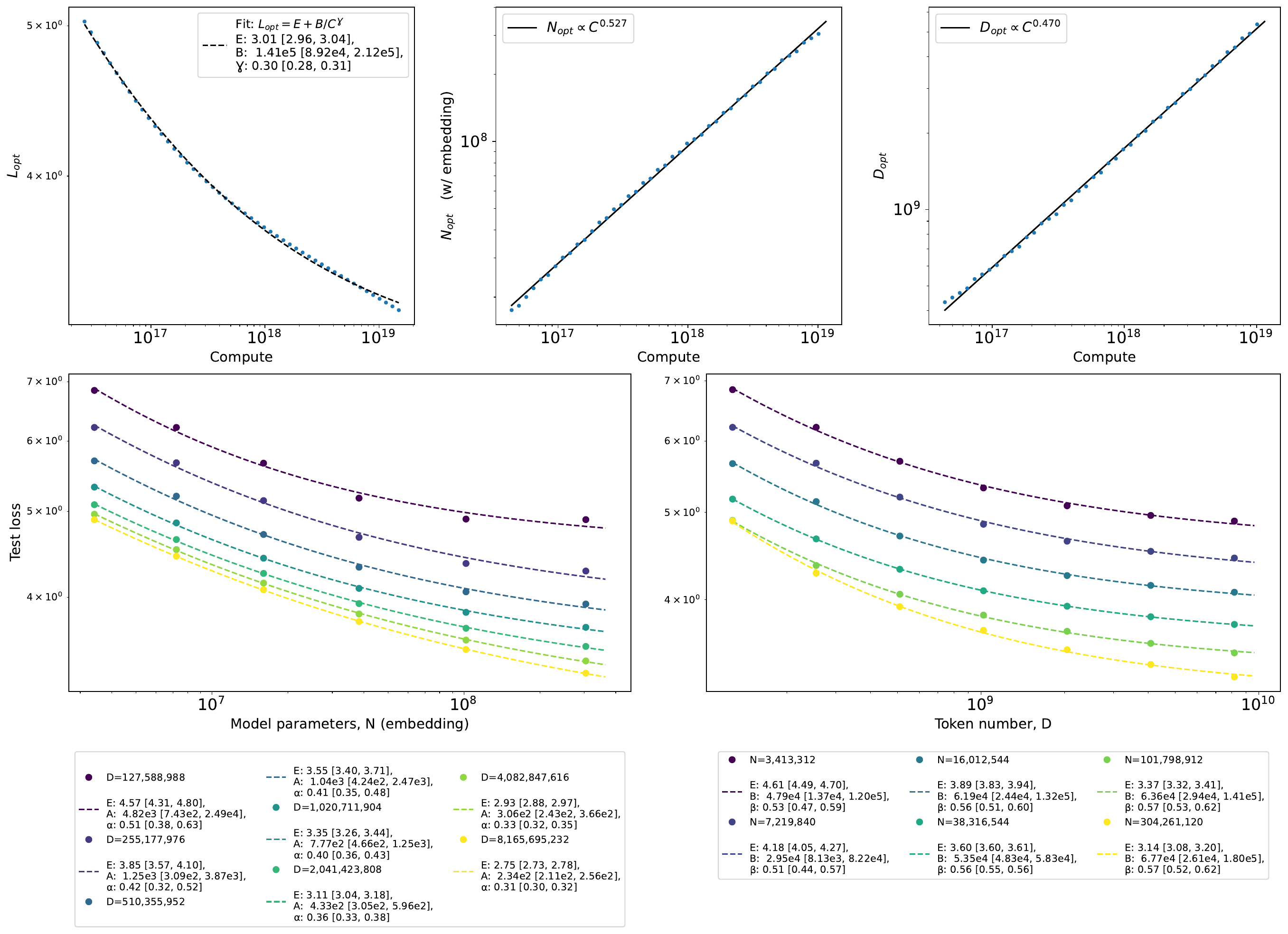}
  \caption{\label{4layermuPlanguageLD_embed}4 layer transformer on language (Fineweb-edu) with Standard Parameterization. Mean exponent $\overline{\alpha_D} = 0.391$ with standard deviation $0.060$. Mean exponent $\overline{\beta_N} = 0.550$ with standard deviation $0.020$. Average MSE for $L(N)_D$ 1d power law fits is $1.28 \times 10^{-3}$, as compared with $7.11 \times 10^{-3}$ for best exponential fit. Average MSE for $L(D)_N$ 1d power law fits is $4.51 \times 10^{-4}$, compared to $7.47 \times 10^{-3}$ for best exponential fit. 
}
  \label{fig:4layer_language_SP}
\end{figure}

\subsection{Barabási-Albert graph with 8K nodes, 50K edges}

\begin{figure}[H]
  \centering
  \includegraphics[width=\linewidth]{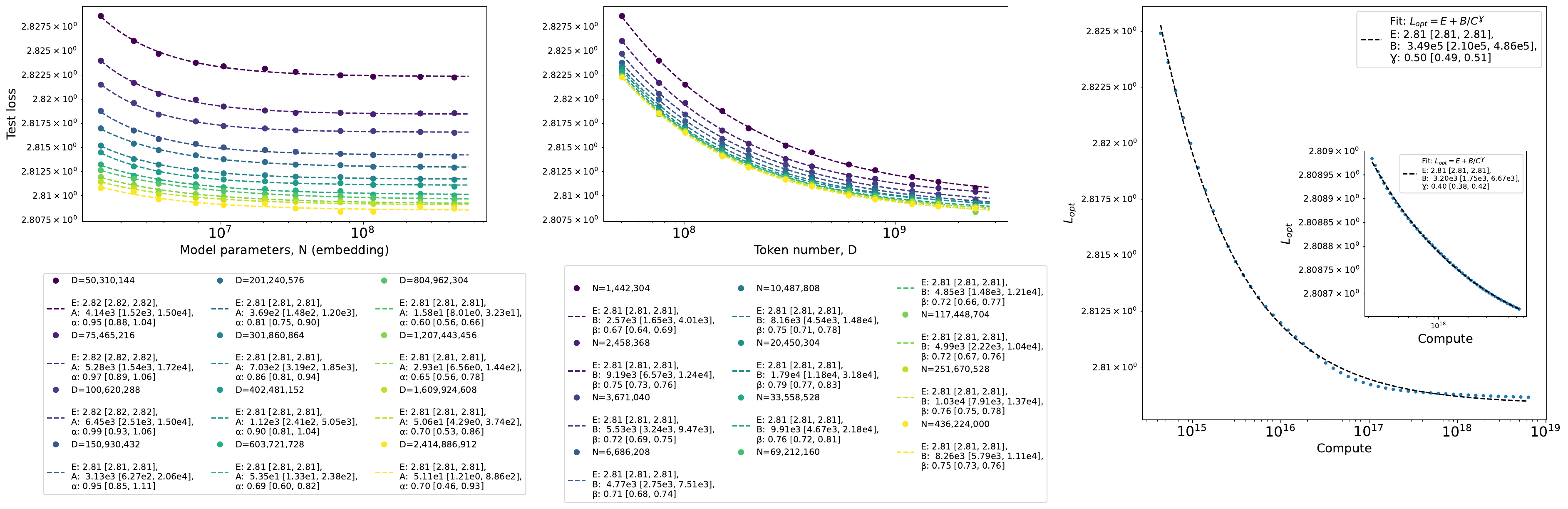}
  \caption{Results for Barabasi-Albert graph with $8,192$ nodes.  Mean $\overline{\alpha_D} = 0.815$ with standard deviation $0.133$. Mean $\overline{\beta_N} = 0.735$ with standard deviation $0.031$.
  Average MSE for $L(N)_D$ 1d power law fits: $1.22\times 10^{-8}$ compared with  $6.29 \times 10^{-8}$ for best exponential fit. Average MSE for $L(D)_N$ 1d power law fits: $1.32\times 10^{-8}$ compared with $5.07 \times 10^{-7}$ for best exponential fit. 
}
  \label{fig:ba}
\end{figure}

Fig. \ref{fig:ba} shows our results for a scale-free Barabási-Albert graph. 

\subsection{Erdös-Renyi graphs with 50K nodes and 2M edges, $\kappa = 0,1$}

\begin{figure}[H]
  \centering
  \includegraphics[width=\linewidth]{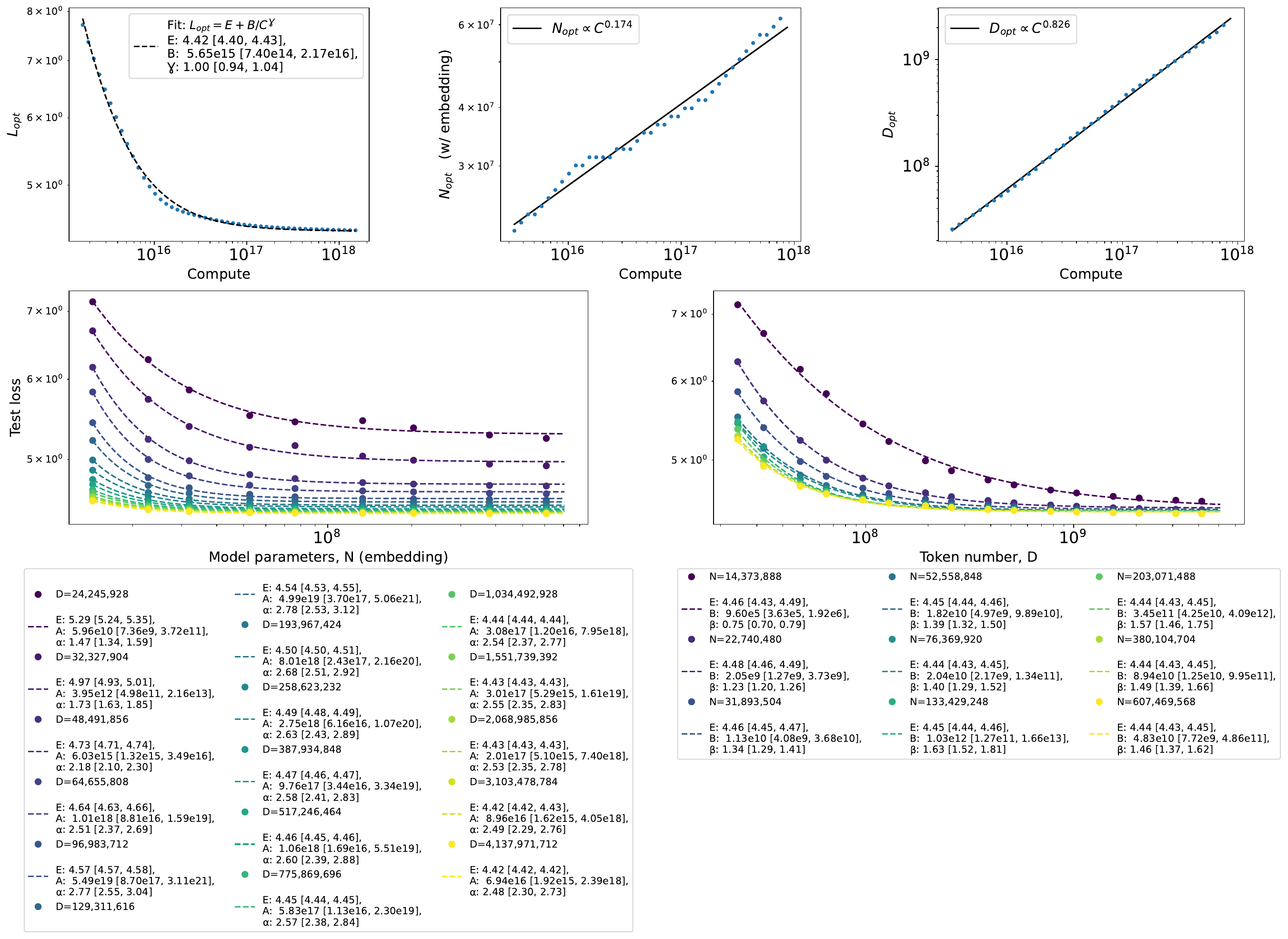}
  \caption{Results for Erdos-Renyi graph with $50,000$ nodes and $2$ million edges, with unbiased sampling $\kappa = 0$. 
  Mean $\alpha = 2.443$, with standard deviation of $0.348$. Mean $\beta = 1.361$ with standard deviation = 0.244. Average MSE for $L(N)_D$ 1d power law fits is $2.76 \times 10^{-4}$ as compared with $8.23 \times 10^{-4}$ for the best exponential fit. Average MSE for the $L(D)_N$ 1d power law fits is $3.99 \times 10^{-4}$, and best exponential fit gives $3.08 \times 10^{-3}$. 
}
  \label{fig:er50k2M_alpha0}
\end{figure}

\begin{figure}[t]
  \centering
  \includegraphics[width=\linewidth]{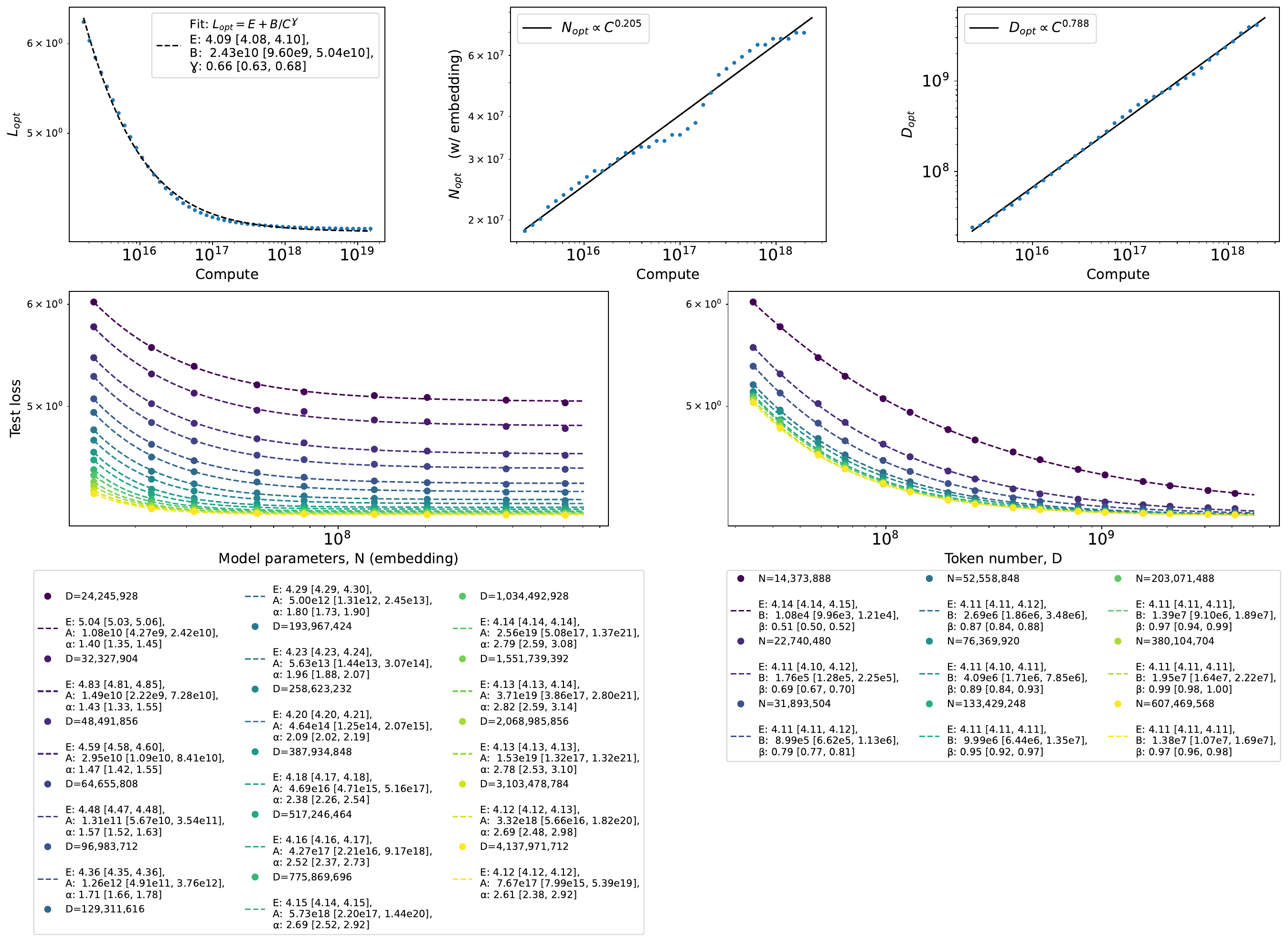}
  \caption{Results for Erdos-Renyi graph with $50,000$ nodes and $2$ million edges, with biased sampling $\kappa = 1$. 
  Mean $\alpha = 2.169$ with standard deviation $0.529$. Mean $\beta = 0.847$ with standard deviation $0.151$. Average MSE for $L(N)_D$ 1d power law fits: $5.30 \times 10^{-5}$, compared with $4.25 \times 10^{-4}$ for best exponential fit. Average MSE for $L(D)_N$ 1d power law fits: $4.05 \times 10^{-5}$, compared with $2.99 \times 10^{-3}$ for best exponential fit. 
}
  \label{fig:er50k2M_alpha1}
\end{figure}

\clearpage
\newpage
\bibliographystyle{assets/plainnat}
\bibliography{paper}

\end{document}